%% file: main.tex
\documentclass[twoside]{article}

\usepackage[accepted]{aistats2022}
\usepackage[round]{natbib}

\usepackage[T1]{fontenc}    
\usepackage[colorlinks=true,citecolor=blue]{hyperref}       
\usepackage{url}            
\usepackage{booktabs}       
\usepackage{amsfonts}       
\usepackage{nicefrac}       
\usepackage{microtype}      
\usepackage{xcolor}         
\usepackage{expdlist}       

\usepackage{amssymb,amsmath}
\usepackage{amsthm}
\usepackage{cleveref}
\usepackage{bm}
\usepackage{mathtools}

\newtheorem{theorem}{Theorem}

\newtheorem{lemma}{Lemma}
\newtheorem{definition}{Definition}

\newcommand{\E}{\mathbb{E}}

\newcommand{\erf}{\operatorname{erf}}
\newcommand{\sign}{\operatorname{sign}}
\newcommand{\smallkl}{\operatorname{kl}}
\newcommand{\KL}[1]{\operatorname{KL}\left(#1\right)} \renewcommand{\Pr}{\mathbb{P}}
\renewcommand{\Re}{\mathbb{R}}
\renewcommand{\vec}[1]{\boldsymbol{#1}}

\DeclareMathOperator{\AV}{AV}

\DeclarePairedDelimiter\ceil{\lceil}{\rceil}

\begin{document}

%

%
%

\twocolumn[

\aistatstitle{On Margins and Derandomisation in PAC-Bayes}

\aistatsauthor{ Felix Biggs \And Benjamin Guedj}

\aistatsaddress{ Centre for Artificial Intelligence \\ Department of Computer Science \\ University College London \\ United Kingdom \And  Centre for Artifical Intelligence \\ Department of Computer Science \\ University College London and Inria \\ United Kingdom and France } ]

\input{body.tex}

\subsubsection*{Acknowledgements}

We warmly thank all reviewers and the area chair for their insightful feedback which improved the quality of our presentation.
Felix Biggs gratefully acknowledges the support of the CDT for Foundational Artificial Intelligence through UKRI grant EP/S021566/1.
Benjamin Guedj acknowledges partial support by the U.S. Army Research Laboratory and the U.S. Army Research Office, and by the U.K. Ministry of Defence and the U.K. Engineering and Physical Sciences Research Council (EPSRC) under grant number EP/R013616/1; Benjamin Guedj also acknowledges partial support from the French National Agency for Research, grants ANR-18-CE40-0016-01 and ANR- 18-CE23-0015-02.

\bibliographystyle{plainnat}
\bibliography{bibliography}


\clearpage
\appendix

\thispagestyle{empty}

\onecolumn \makesupplementtitle

\input{appendix.tex}

\end{document}

%% file: body.tex

\begin{abstract}
  We give a general recipe for derandomising PAC-Bayesian bounds using margins,
  with the critical ingredient being that our randomised predictions concentrate
  around some value. The tools we develop straightforwardly lead to margin
  bounds for various classifiers, including linear prediction---a class that
  includes boosting and the support vector machine---single-hidden-layer
  neural networks with an unusual \(\erf\) activation function, and deep ReLU
  networks. Further, we extend to partially-derandomised predictors where only
  some of the randomness is removed, letting us extend bounds
  to cases where the concentration properties of our predictors are otherwise
  poor.
\end{abstract}

\section{INTRODUCTION}\label{section:introduction}

PAC-Bayesian\footnote{PAC-Bayes theory originates in the seminal papers from
  \citet{STW1997}, \citet{McAllester1998} and \citet{McAllester1999}, and
  was further formalised by \citet{catoni2007}, among others -- we refer to
  \citet{guedj2019primer} for a recent overview.} generalisation bounds have
recently seen a resurgence of interest after the comparative successes of a
series of papers applying them to deep neural networks, beginning with
\citet{dziugaite2017computing,DBLP:conf/nips/Dziugaite018,DBLP:conf/iclr/NeyshaburBS18},
and \citet{NIPS2019_8911,DBLP:conf/iclr/ZhouVAAO19}. One can use these to
understand where to apply techniques and motivate new learning algorithms
\citep[as for example][]{DBLP:conf/iclr/ForetKMN21}, as well as provide
certification for a given predictor and address the more ambitious goal of
understanding generalisation.

One particularly useful aspect of PAC-Bayesian results compared to standard
PAC/VC results is that they are non-uniform: the tightness of the guarantee on
the generalisation error depends on the specific predictor chosen, not merely
on its performance on the training set. This is necessary in cases where our
broad class can easily overfit---as for example with many neural networks
architectures, which were shown by \citet{DBLP:journals/cacm/ZhangBHRV21} to be
able to fit random training labels---since any guarantee must then selectively
favour predictors which are reasonable given real data. If our strategic choice
of learner turns out to be a good match in practice to the data-generating
distribution, the bound should reflect this.

But how to measure this match based only on the training data? A common approach
is that we should not just take into account the train error of a given
predictor but also its \emph{confidence}. One way to formalise this is the
concept of a margin, introduced to bound the error of the perceptron
\citep{novikoff62convergence} and later used to motivate the support vector
machine \citep{DBLP:journals/ml/CortesV95}. A confident predictor with a large
margin on a given example will be locally robust to parameter (and data)
perturbations. If this is true across the dataset our bounds should reflect this
and be tighter. From the perspective of Occam's razor this robustness leads to a
large set of valid perturbations giving near-equivalent (in terms of dataset
outputs) predictors; some predictor in this set is thus likely to be close
to a ``simple'' prediction rule of the kind that we should generally favour
\citep{schapireBoostingMarginNew1998}.

Remarkably, this idea of parameter robustness---as measured by margins---can be
formalised through the lens of PAC-Bayes, which more typically bounds the loss
of randomised predictors (although there are notable exceptions, see related
work in \Cref{section:derandomising}). After picking a derandomised prediction
rule, we can construct a (weighted) class of ``proxy'' predictors that
approximate this rule, with the size and diversity of this class growing with
the allowed margin. Since a larger such class is more likely to overlap strongly
with our PAC-Bayesian prior, tighter bounds are obtained for larger margins.
This idea has been used informally by \citet{langford01boundsfor} and
\citet{langford2003pac}: here we formalise and extend it considerably.

A critical ingredient in this process is the construction of randomised
classifiers that have favourable concentration properties---simply, that the
parameters are robust to perturbations---so that their deviations from a central
derandomised prediction rule are bounded with high probability. The insight that
these deviations need only be controlled with high probability rather than
certainty is crucial in obtaining better rates and simplifying proofs, and opens
the door to the application of powerful concentration of measure results.

We go further and introduce the idea of partially-derandomised predictors, which
remove only some of the randomness from the proxy predictors. This enables us to
obtain bounds in further cases which would otherwise be difficult to address. We
hope that these tools can be further developed to address situations where
classical bounding techniques have not worked well (such as in deep neural
networks), and that our derived corollaries---such as that for linear
prediction---can be used in practice for the provision of self-certifying
predictors and model selection.

\textbf{Contributions and structure.} In \Cref{section:derandomising} we
discuss and formalise the derandomisation of PAC-Bayesian bounds using margins
and averaging, introduce the slight generalisation of sub-Gaussian random
variables that enables many of our results, and compare to the covering number
(or \(\gamma\)-ball) approach. Corollaries of these methods include margin
bounds for the following:
\begin{description}[\compact\topsep=0em]
  \item[\Cref{section:linear}] \(L_2\) and \(L_1\) normed linear prediction; in the
    \(L_2\) ``hard-margin'' case this improves on the bound of
    \citet{bartlett1998generalization} and matches the lower bound of
    \citet{DBLP:conf/icml/GronlundKL20}; the other bounds match the state of the art with simpler proofs.
  \item[\Cref{section:partial-agg}] linear predictors with a randomised (and potentially learned) feature space.
  \item[\Cref{section:one-hidden}] one-hidden-layer neural networks with \(\erf\) activations. This involves an interesting new randomised predictor taking the form of a mixture distribution. The introduction of partial-derandomisation also enables bounds with the final two layers derandomised, but the initial layers having randomised weights.
  \item[\Cref{section:beyond-two-layers}] deep ReLU networks, a slight improvement on and with proof ideas drawing from that given by \citet{DBLP:conf/iclr/NeyshaburBS18}.
\end{description}
Finally, in \Cref{section:conclusion}, we summarise and provide an overview of
our results. Related work for specific applications is discussed in the
corresponding section, with general work on derandomising bounds in
\Cref{section:derandomising}.

\textbf{Notation.} We will consider classification of i.i.d. examples from a
distribution, \(\mathcal{D}\), on some product space
\(\mathcal{Z} = \mathcal{X} \times \mathcal{Y}\), by vector-valued predictors in
a function space \(\mathcal{H} \subset \widehat{\mathcal{Y}}^{\mathcal{X}}\). For
binary classification \(\mathcal{Y} = \{+1, -1\}, \; \widehat{\mathcal{Y}} = \Re\)
and we take the sign of the output as our prediction, while for multi-class
prediction,
\(\mathcal{Y} = [c] := \{1, \dots, c\},\; \widehat{\mathcal{Y}} = \Re^{c}\) and the
maximum argument is the prediction.

The multi-class margin, \(M: \mathcal{H} \times \mathcal{Z} \to \Re\) is the mapping
\[ M(f, (x, y)) := f(\vec{x})[y] - \max_{y' \ne y}f(\vec{x})[y'] \] where by
\(f(x)[y]\) we indicate the \(y\)th component of \(f(x)\). In a slight abuse of
notation we also define the binary margin \(M(f, (x, y)) := yf(x)\).

We define the margin error
\(L_{\gamma}(f) := \Pr_{z \sim \mathcal{D}}\{M(f, z) \le \gamma\}\), also
writing \(L(f) := L_0(f)\) for the misclassification loss or probability of
error, and \(\hat{L}_{\gamma}(h) := m^{-1}|\{(x, y) \in S : yh(x) < \gamma\}|\)
for the empirical margin error (defined for some sample
\(S \sim \mathcal{D}^{m}\) and margin \(\gamma \ge 0\)). We will also use the
abbreviation \(\Pr_{\mathcal{D}}(A) = \Pr_{z \sim \mathcal{D}}(A)\) and similar
for the expectation.

As is common in the PAC-Bayes literature, when considering distributions over
predictors \(P \in \mathcal{M}_1(\mathcal{H})\) (by which we denote the space of
probability measures on \(\mathcal{H}\)), we write
\(L(P) := \mathbb{E}_{f \sim P} L(f)\) interchangeably with the above, and
analogously with other margin errors\footnote{When considering functions in
\(\mathcal{H}\) parameterised by some space \(\Theta\) (of which
\(\mathcal{H}\) may be a quotient), we will be somewhat loose with
interchanging \(\mathcal{M}_1(\mathcal{H})\) with \(\mathcal{M}_1(\Theta)\);
this will not affect any results in practice, as the Kullback-Liebler divergence between distributions on the \(\Theta\) upper bounds that of their image distributions on \(\mathcal{H}\).}.

\section{PAC-BAYES FOR APPROXIMATIONS}\label{section:derandomising}

In \Cref{subsection:approximation} we discuss a measure of difference when
substituting or ``approximating'' one distribution over prediction functions by
another through a coupling method, and from this derive (partially) derandomised
PAC-Bayesian margin bounds in \Cref{subsection:bounds}. In
\Cref{subsection:subgaussian} we discuss a powerful concentration-based method
for constructing approximating distributions, and in \Cref{subsection:covering}
relate our techniques to covering number methods.

\textbf{Technical overview.}
PAC-Bayesian analysis makes a different set of assumptions to Bayesian analysis;
the guarantees obtained are simultaneously true for any 
data-generating
distribution and PAC-Bayes prior, \(P_0\). This prior serves as a reference
measure for the PAC-Bayes posterior \(P\), in a Kullback-Liebler divergence
complexity term, \(\KL{P, P_0}\). Poor choice of \(P_0\), or an
overly-concentrated \(P\) will lead to valid but trivial bounds.

If we wish to use a predictive distribution \(Q\) concentrated on a single
predictor \(h\), we must replace the loss terms by approximating them by a
class, \(\mathcal{P}\), of randomised ``proxy'' functions, each of which has
margins differing by less than \(\gamma > 0\) with high probability
\(\ge 1 - \epsilon\) from \(h\). The PAC-Bayes bounds then obtained can use the
minimising proxy from the prior,
\(\kappa = \min_{P \in \mathcal{P}} \KL{P, P_0}\), leading to bounds of the
overall form: with probability greater than $1-\delta$, 
\[ L(Q) \le O\left(\hat{L}_{\gamma}(Q) + \epsilon + \frac{\kappa(Q, P_0; \epsilon, \gamma) + \log(1/\delta)}{m} \right).\]
This \(\epsilon\) high probability term (which we will usually set
\(\epsilon \in O(m^{-1})\)) is a crucial difference from the covering number approach, and
considerably simplifies proofs, since for any proxy \(P \in \mathcal{P}\) we
only need concentration-of-measure (in the margins) around \(h\). A second
innovation is to leave open the possibility of using a partially-derandomised
predictive distribution, \(Q\), through a coupling method.

\textbf{Related work.}
Work on derandomised PAC-Bayesian bounds is as old as the field; particularly
when relating to the average prediction. Bounds holding with high probability
over a sampled predictor (directly drawn from the PAC-Bayes posterior) appear \emph{e.g.} in \citet{catoni2007}, \cite{AB2013,guedj2013}.

\subsection{Approximation of predictive distributions}\label{subsection:approximation}

Let \(P, Q \in \mathcal{M}_1(\mathcal{H})\) be distributions on prediction
functions in \(\mathcal{H}\), as generally considered in PAC-Bayes. We denote by
\(\Pi(P, Q) \subset \mathcal{M}_1(\mathcal{H} \times \mathcal{H})\) the set of
product distributions with marginals \(P\) and \(Q\) (also known as couplings
between \(P\) and \(Q\)). For each of these distributions, the margins of these
functions are sets of real variables indexed by \(\mathcal{Z}\) (and can
equivalently be viewed as real-valued stochastic processes on \(\mathcal{Z}\)).

We define the upper \(\gamma\)-approximate variation of these margins (defined
as a relaxation of the total variation distance) using the relative margin at z,
\(d_{z}(f, g) = M(f, z) - M(g, z)\) as (abbreviating \(\Pr_{(f, g) \sim \pi}\)
as \(\Pr_{\pi}\) here and henceforth, with a similar notation for the
expectation),
\begin{equation*}
\text{UAV}_{\gamma}(P, Q) := \inf_{\pi \in \Pi(P, Q)} \sup_{z \in \mathcal{Z}} \Pr_{\pi}\left\{ d_z(f, g) > \gamma/2 \right\}.
\end{equation*}

In the case of the binary margin, \(M(f, (x, y)) = yf(x)\) with
\(y \in \{+1, -1\}\), so the above is symmetric under interchange of \(P\) and
\(Q\); this is not true in general. We therefore define the symmetrised version,
the \(\gamma\)-approximate variation on \(\mathcal{Z}\), as
\begin{equation*}
\text{AV}_{\gamma}(P, Q) := \max(\text{UAV}_{\gamma}(P, Q), \text{UAV}_{\gamma}(Q, P)).
\end{equation*}

We say \(P\) and \(Q\) \((\gamma, \epsilon)\)-approximate each other on
\(\mathcal{Z}\) if \(\text{AV}_{\gamma}(P, Q) \le \epsilon\). This is a
margin-based generalisation of the total variation distance: the zero-margin
approximate variation, \(\AV_{0}(P, Q)\), is equal to the total variation
distance between the distributions on margins,
\(\delta_{\text{TV}}(M(P, z), M(Q, z)) \le \delta_{\text{TV}}(P, Q)\). However
the total variation distance is too strict to yield non-vacuous bounds in most
cases where \(P \ne Q\). Approximation implies the possibility of substituting
one margin loss for another at the cost of these terms and a margin.

\begin{lemma}\label{lemma:margin_subsitution}
Let \(\gamma > 0, \epsilon \ge 0\); if \(P, Q \in \mathcal{M}_1(\mathcal{H})\) with
\(\text{AV}_{\gamma}(P, Q) \le \epsilon\), then for any data distribution
\(\mathcal{D}\), \(L(P) \le L_{\gamma/2}(Q) + \epsilon\) and
\(L_{\gamma/2}(Q) \le L_{\gamma}(P) + \epsilon\).
\end{lemma}

\begin{proof}
For any events \(A, B\), \(\Pr(A) \le \Pr(B) + \Pr(\bar{B} \cap A)\); and for any
coupling \(\pi \in \Pi(P, Q)\) we have
\begin{align*}
L(P) &= \E_{f \sim P}\Pr_{\mathcal{D}}\left\{ M(f, z)  \le 0 \right\} \\
&\le \E_{g \sim Q} \Pr_{\mathcal{D}}\left\{ M(g, z)  \le \gamma/2 \right\} \\
& \: \:+\: \E_{\pi}\Pr_{\mathcal{D}} \left\{ M(g, z)  > \gamma/2 \, \land \, M(f, z) \le 0 \right\} \\
&\le L_{\gamma/2}(Q) \:+\: \E_{\mathcal{D}} \Pr_{\pi}\left\{ d_z(f, g) > \gamma/2 \right\}.
\end{align*}
Replacing the expectation with its pointwise bound and taking the infimum over
couplings, we find that
\(L(P) \le L_{\gamma/2}(Q) + \text{UAV}_{\gamma}(P, Q)\). An analogous process
follows for the other side, with the order of \(Q\) and \(P\) reversed.
\end{proof}

\textbf{Remark.}
A less sophisticated analysis could have used the bound
\(|M(f, z) - M(g, z)| \le 2 \max_{y \in \mathcal{Y}}|f(x)[y] - g(x)[y]|\)
instead, leading to similar PAC-Bayes bounds. This definition of AV improves
constants in some derived bounds and removes a factor of \(c\), the number of
classes. We also note that the coupling need not be the same on both sides of
the bounds, although we do not use this in later proofs.

\subsection{PAC-Bayes bounds with approximations}\label{subsection:bounds}

\Cref{lemma:margin_subsitution} can be used to derive a type of PAC-Bayesian
bound for a predictive distribution \(Q\), as follows. First we define the
\((\gamma, \epsilon)\)-approximating KL projection onto a prior
\(P_0 \in \mathcal{M}_1(\mathcal{H})\) (defined independently of the data), of
\((\gamma, \epsilon)\)-approximations to \(Q\).

\begin{definition}\label{constrained_complexity}
\emph{Approximating KL-Projection}: Given some prior distribution \(P_{0}\) on
\(\mathcal{H}\), the \((\gamma, \epsilon)\)-approximate projection of \(Q\)
onto \(P_0\) is
\begin{equation*}
\kappa(Q, P_0; \gamma, \epsilon) := \min_{P \in \mathcal{P}} \operatorname{KL}(P, P_{0})
\end{equation*}
where
\(\mathcal{P} = \{ P \in \mathcal{M}^{1}(\mathcal{H}), ~  \operatorname{AV}_{\gamma}(P, Q) \le \epsilon \}\).
\end{definition}

This can be viewed as the ``closest'' (in the KL sense) \emph{proxy} \(P\), to
our prior that approximates \(Q\) sufficiently well. In practice, we will
restrict the family of proxies, \(\mathcal{P}\), to some more tractable set and
construct \(P\) explicitly. The notion can be used in combination with many
PAC-Bayesian bounds, replacing the usual KL divergence with a new complexity
term and the losses with margin losses. We give the following formulations as
examples.

\begin{theorem}\label{tightened_version}
  Given data distribution \(\mathcal{D}\) on \(\mathcal{X} \times \mathcal{Y}\),
  prior \(P_{0} \in \mathcal{M}^{1}(\mathcal{H})\),
  \(\gamma > 0, \epsilon \in [0, \frac{1}{2}]\) and
  \(\delta \in (0, 1)\), the following hold each with probability
  \(\ge 1 - \delta\) over \(S \sim D^{m}\), for all
  \(Q \in \mathcal{M}^{1}(\mathcal{H})\)
  \begin{description}
    \item[``small-kl'']
    \begin{equation*}\label{actual_bound_small_kl}
    \smallkl(\hat{L}_{\gamma}(Q) + \epsilon : L(Q) - \epsilon )
    \le \frac{\kappa(Q, P_{0}; \gamma, \epsilon) + \log\frac{2\sqrt{m}}{\delta}}{m} ,
    \end{equation*}

    \item[``interpolating''] (given that \(\hat{L}_{\gamma}(Q) = 0\))
    \begin{equation*}\label{fast_rate}
    L(Q) \le \frac{\kappa(Q, P_{0}; \gamma, \epsilon) + \log\frac{1}{\delta}}{m} + 4\epsilon \log\frac{1}{\epsilon},
    \end{equation*}
  \end{description}
  where \(\smallkl(q : p) = q\log\frac{q}{p} + (1-q)\log\frac{1-q}{1-p}\) if
  \(p \ge q\) and otherwise \(0\) (this formulation is monotonic in \(q\) and thus
  one-sided).
\end{theorem}

\begin{proof}
The standard PAC-Bayesian bounds (we refer to \Cref{pac-bayes-theorems-reference}) with loss function
\(\ell_{\gamma/2}\) are true for the minimising \(P\) in
\Cref{constrained_complexity}. We can then use \Cref{lemma:margin_subsitution}
to replace the losses with those w.r.t. \(Q\).

In the final step, we use the following formulation of the Catoni bound
using \citet[Proposition 2.1]{germainPACBayesianLearningLinear2009}, valid if \(\epsilon\) is independent of
the sample \(S\) and \(\hat{L}_{\gamma}(Q) = 0\):
\begin{equation*}
\smallkl(\epsilon : L(Q) - \epsilon )
\le \frac{1}{m} \left( \KL{P, P_0} + \log\frac{1}{\delta} \right)
\end{equation*}
and adapt it in the same way as the previous bounds to \(\kappa\). We then use
the lower bound
\(\smallkl(\epsilon : p - \epsilon) \ge p + 4\epsilon \log \epsilon\), valid for
all \(\epsilon \in [0, \frac{1}{2}], p \in [0, 1]\). This is proved in
\Cref{lemma:small-kl-lower-bound} in supplementary material.
\end{proof}

\subsection{Sub-Gaussian derandomisation}\label{subsection:subgaussian}

One simple case to which the above bounds can be applied, often in a
dimension-independent way, is that of total derandomisation by \emph{averaging}:
for some \(P\), we set \(Q = \delta(F_P)\), a point mass measure on the
\(P\)-aggregate function \(F_P(x) := \E_{f \sim P}f(x)\). If the score does not
vary too much under \(P\), as defined by a sub-Gaussian condition, derandomised
PAC-Bayes bounds follow directly through our framework. This is a formalisation
of a proof idea, used for example by \citet{langford2003pac}; by making this
connection explicit we contribute a clearer understanding of PAC-Bayes derived
margin bounds.

First we define the idea of sub-Gaussian random functions, defined here in a
slightly more general way to accommodate ``partial-derandomisation''.

\begin{definition}
We say a coupling \(\pi \in \Pi(P, Q)\) is \(\sigma^2\)-sub-Gaussian on
\(\mathcal{Z}\) if
\[\E_{\pi} \exp(t (f(x)[y] - g(x)[y])) \le \exp(t^{2}\sigma^{2}/2)\]
and \(E_{f \sim P} f(x)[y] = E_{g \sim Q} g(x)[y]\), for all \(t \in \Re\),
\((x, y) \in \mathcal{Z}\). The square bracket indicates the \(y\)th index if
the output is multi-dimensional; in the scalar case we remove it.
\end{definition}

We will further stretch this definition and call a \emph{single} distribution,
\(P\), \(\sigma^2\)-sub-Gaussian, if the trivial coupling
\(\pi = P \otimes \delta(F_P)\) is. Sub-Gaussianity implies bounds on the
approximate variation:

\begin{lemma}\label{sub-gaussian-lemma}
If \(\pi \in \Pi(P, Q)\) is \(\sigma^2\)-sub-Gaussian on
\(\mathcal{Z} = \mathcal{X} \times \mathcal{Y}\), for binary and multi-class classification respectively,
\begin{equation*}
\operatorname{AV}_{\gamma}(P, Q) \le
\begin{cases}
\exp(-\gamma^2/8\sigma^2) & \text{for  } \mathcal{Y} = \Re, \\
\exp(-\gamma^2/16\sigma^2) & \mathcal{Y} = \Re^{c}. \\
\end{cases}
\end{equation*}
\end{lemma}

\begin{proof}
  Considering the zero-mean random variable \(X = f(x)[y] - g(x)[y]\) for
  \((f, g) \sim \pi\) (\(\sigma^2\)-sub-Gaussian) and fixed
  \((x, y) \in \mathcal{Z}\), the Chernoff bound \citep[see][for example, for a
  thorough introduction to sub-Gaussianity]{DBLP:books/daglib/0035704},
  immediately implies
\[\Pr(X > t) \lor \Pr(-X > t) \le e^{t^2/2\sigma^2}\] for all \(t > 0\).
In the binary margin case, \(M(f, z) = yf(x)\) which is either \(f(x)\) or
\(-f(x)\); setting \(t = \gamma / 2\) in the above therefore gives the bound.

In the multi-class case we consider the upper bound obtained by letting \(y'\)
achieve the maximum margin for \(g\); then \(M(f, z) \le f(x)[y] - f(x)[y']\),
so
\begin{align*}
&\Pr_{\pi}\left\{ M(f, z)  - M(g, z) > \frac{\gamma}{2} \right\} \\
&\le \Pr_{\pi}\left\{ f(x)[y] - f(x)[y'] -  g(x)[y] + g(x)[y'] > \frac{\gamma}{2} \right\}.
\end{align*}
Since both \(f(x)[y] - g(x)[y]\) and \(f(x)[y'] - g(x)[y']\) are
\(\sigma^2\)-sub-Gaussian, their sum is \(2\sigma^2\)-sub-Gaussian and the bound
follows by repeating the process on with signs reversed.
\end{proof}

\textbf{Other concentration assumptions.} We note that although we only discuss
sub-Gaussian concentration, it is possible to require other
concentration properties, for example sub-exponential ones; our framework easily
accommodates these. Sub-Gaussianity is only the simplest way to ensure such
concentration, and we primarily consider it in our later results as it already
leads to simple proofs of margin bounds in multiple settings.

\subsection{Relation to covering}\label{subsection:covering}

Here we discuss how our bounds can be used to derive a standard covering
approach as a sub-case; we show that this leads to certain problems, which are
circumvented by the concentration approach. A further consequence is that
covering-based bounds usually lead to ``uniform'' bounds which are subject to
problems discussed in \citet{DBLP:conf/nips/NagarajanK19}. All the bounds we
provide in later sections are non-uniform and avoid these pitfalls.

By setting \(\epsilon = 0\) with certain choices of prior we can obtain a fairly
standard ``covering'' approach: call \(N_{\gamma}\) a \(\gamma\)-net of
\(\mathcal{H}\), if for any \(f \in \mathcal{H}\), there exists
\(g \in N_{\gamma}\) such that \(|M(f, z) - M(g, z)| \le \gamma\) for all
\(z \in \mathcal{Z}\). If we choose a prior supported everywhere on a
\(\gamma/2\)-net for \(\mathcal{H}\), we can achieve \(\AV_{\gamma}(P, Q) = 0\)
for any \(Q \in \mathcal{M}_1(\mathcal{H})\), including \(Q\) supported on just
a single predictor. The simplest approach chooses \(P_0\) as uniform on these
points so that
\[ \kappa(Q, P_0; \gamma, 0) \le \log |N_{\gamma/2}| \] where \(|N_{\gamma/2}|\)
is the cardinality of the net. A more sophisticated choice of non-uniform prior
enables structural risk minimisation-type covering number bounds.

However, such bounds will typically be dependent on the dimension of the
parameter space, as demonstrated by the following proposition (proved in
\Cref{section:cover-lower-bound}) for linear classification. Our bounds in the
following sections will avoid this dimension-dependence.

\begin{theorem}\label{theorem:cover-lower-bound}
  Consider binary classification with functions \(f_w = \langle w, x \rangle\)
  and \(x \in \Re^d, \|x\|_2 \le 1\). For any prior \(P_0\) on weights in
  \(\Re^d\), there exists a prediction distribution \(Q\) supported on
  \(\|w\|_2 \le 1\) such that
  \[ \kappa(Q, P_0; \gamma, \epsilon=0) \ge \Omega(d).\]
\end{theorem}

\section{LINEAR PREDICTION BOUND}\label{section:linear}

Here we demonstrate our framework in action by deriving generalisation bounds
for linear predictors. These bounds essentially follow from an initial Gaussian
assumption combined with the sharp (sub-Gaussian) concentration of the predictor
output around its mean. We hope they can be useful for self-certification in the
low data regime, and for model (or kernel) selection without a validation set.

\textbf{\(L_2\)-normed linear predictors.} This situation has been considered by
a large number of papers, from \citet[Theorem 1.7, using
Fat-Shattering]{bartlett1998generalization} in the fast-rate or interpolating
case, to \citet[Theorem 22, using Rademacher complexity]{bartlett2002rademacher}
in the ``soft-margin'' case. \citet{DBLP:conf/colt/McAllester03} presents
alternative bounds in the ``soft-margin'' case, and is itself an attempt to find
a expression for the implicit PAC-Bayesian result of \citet{langford2003pac}. We
give bounds for both cases, through a proof similar to the method of
\citet{langford2003pac}, but using a different base PAC-Bayesian bound which
makes solving the interpolating hard-margin scenario of
\citet{bartlett1998generalization} more straightforward.

\begin{theorem}\label{svm_bound}
  In the binary classification setting with \(\mathcal{X}\) a Hilbert space with
  \(\|x\|_{2} \le R\), and \(\delta \in (0, 1)\), with probability
  \(\ge 1 - \delta\) over \(S \sim D^{m}\), for all linear predictors
  \(F_w(x) = \langle w, x\rangle \) with \(\|w\|_2 \le 1\) and all
  \(\gamma > 0\) (``soft-margin''),
\[ L(F_{\vec{w}}) \le \hat{L}_{\gamma}(F_{\vec{w}}) + \sqrt{\frac{\hat{L}_{\gamma}(F_{\vec{w}}) \cdot \Delta}{m}} + \frac{\Delta + \sqrt{\Delta} + 2}{m}, \]
where we define \(\Delta := 2\log(2/\delta) + 9(R/\gamma)^2\log m\). Additionally, under the
same conditions and probability, provided
\(\gamma_{\star} = \max\{\gamma > 0 : \hat{L}_{\gamma}(F_w) = 0\}\) exists (``hard-margin''),
\begin{equation*}
L(F_w) \le \frac{8 (R/\gamma_{\star})^2\log m + \log (1/\delta)}{m}.
\end{equation*}
\end{theorem}

\begin{proof}
Without loss of generality assume \(R = 1\). To consider a free choice of
margin \(\gamma\), we note the scaling property
\(\bm{1}\{M(F_{\vec{w}}, z) \le \gamma\} = \bm{1}\{M(F_{(\theta/\gamma) \vec{w}}, z) \le \theta\}\).
This suggests approximating the mean predictor \(F_{(\theta/\gamma) \vec{w}}\)
by the distribution \(P\) over functions
\(f = \langle \vec{u}, \vec{x} \rangle\) for
\(\vec{u} \sim \mathcal{N}((\theta/\gamma) \vec{w}, \bm{I})\). Choosing a
data-free prior \(P_0\) of a similar form, but with
\(\vec{u} \sim \mathcal{N}(\vec{0}, \bm{I})\) gives a divergence
\(\KL{P, P_0} = \frac{1}{2} \|(\theta/\gamma) \vec{w}\|^2 = \theta^2/2\gamma^2\).

\(P\) is \(1\)-sub-Gaussian, so by \Cref{sub-gaussian-lemma},
\(\operatorname{AV}_\theta(P, \delta(F_{(\theta/\gamma) \vec{w}})) \le \exp(-\theta^2/8) = \epsilon\).
Plugging into (the hard-margin) \Cref{tightened_version} we obtain for a fixed
\(\theta^2 = 8\log m \) and all \(\gamma_{\star}\) such that
\(\hat{L}_{\gamma_{\star}}(F_{\vec{w}}) = 0\),
\begin{align*}
L(F_{\vec{w}})
& \le \frac{\theta^2/2\gamma_{\star}^2 + \log\frac{1}{\delta}}{m} + \frac{1}{2}\theta^2 \exp(-\theta^2/8) \\
& \le \frac{4(1 + 1/\gamma_{\star}^2)\log m + \log\frac{1}{\delta}}{m}.
\end{align*}
By the assumptions on \(\|\vec{w}\|_2\) and \(R\), we have
\(\gamma_{\star} \le 1\) to prove the second statement.

Repeating the above but replacing the use of the hard-margin bound with the
small-kl formulation in \Cref{tightened_version} (and using the vacuity of the
bound when \(\gamma > 1\)), we have the tight bound
\begin{multline}\label{equation:smallkl-svm}
  \smallkl(\hat{L}_{\gamma} + m^{-1} : L - m^{-1} ) \\
           \le \frac{1}{m} \left(4(R/\gamma)^2 \log m + \frac{1}{2}\log m + \log \frac{2}{\delta}\right) \le \frac{\Delta}{2m}
\end{multline}
with probability \(\ge 1 - \delta\). To relax the above we use the lower bound
\(\smallkl(q : p) \ge (p-q)^2/(2p)\) for \(p > q\) from
\citet{DBLP:conf/colt/McAllester03} to show
\(L \le \hat{L}_{\gamma} + 2m^{-1} + \sqrt{(\hat{L}_{\gamma} + m^{-1}) \cdot \Delta / m} + \Delta/m\)
which completes the proof.
\end{proof}

In the ``hard margin'' case \Cref{svm_bound} improves on
\citet{bartlett1998generalization} by a factor of \(O(\log m)\), matching the
lower bound of \citet[Theorem 4]{DBLP:conf/icml/GronlundKL20}. In the
``soft-margin'' case \Cref{svm_bound} is of the same order as the state-of-the-art
bound given by \citet{DBLP:conf/icml/GronlundKL20} but with explicitly stated
constants.
We emphasise the extreme simplicity of our proof compared to that given for
Theorem 2 in \citet{DBLP:conf/icml/GronlundKL20} (in Section 2,
p.3-7), and that these are the tightest explicitly-stated bounds for the problem
to our knowledge.

Our soft-margin result in \Cref{svm_bound} also improves upon
\citet{DBLP:conf/colt/McAllester03}, which is proved via a similar method, but
using a different base PAC-Bayes bound that leads to a weaker result. We present
our soft-margin result in the more straightforward form given by the papers
above so these comparisons can be easily made, even though
\Cref{equation:smallkl-svm} is technically tighter. We discuss these differences
at length and give full forms of existing results in \Cref{section:comparison}.

We note that the soft-margin formulation of the bound is true universally
across \(\gamma > 0\), allowing the bound can be optimised for \(\gamma\) in
\(O(m)\) time. If the margin is large for most examples, we can choose
\(\gamma\) so that \(\hat{L}_{\gamma}\) is small and thus the \(\Delta/m\) term
(which is of the same order as the hard-margin bound) dominates. Since the
minimum margin can be sensitive to outliers, this bound will often be tighter
than the hard-margin one.

The margin only appears in the bounds of \Cref{svm_bound} in a ``normalised''
form, \(\gamma/R\), otherwise scaling the data would affect the bound. However,
we note that the bound can sometimes be decreased by normalising the data (as
this maximises the margin for every data point), so we recommend this when using
such predictors.

Finally we also acknowledge the result of \citet[Theorem 1]{DBLP:conf/alt/HannekeK21}
which gives a \emph{algorithm}-dependent hard-margin bound specifically for the
SVM output, and eliminates the \(\log m\) factor. This is provably optimal in
the algorithm-dependent (as ours is in the general) case, which is shown in
\citet[Theorem 5]{DBLP:conf/icml/GronlundKL20}.

\textbf{\(L_1/L_{\infty}\)-normed linear predictors.} \Cref{svm_bound} is a
bound under \(L_2\) norms for \(\mathcal{X}\) and \(w\), applying to situations
such as the SVM. For completeness we provide in \Cref{section:l-infty-proof} a
bound for linear classification under different norm constraints, where the
\(L_1\) norm of the weights and \(L_{\infty}\) norm of the features is
restricted, as in boosting.

These results are essentially the same as the k-th margin bound of
\citet{DBLP:journals/ai/GaoZ13a}, or the central result of
\citet{langford01boundsfor}, but proved through our framework instead, which we
believe provides a unifying perspective. The fundamental proof idea is to
approximate our predictor by a randomised, unweighted, sum of features, as
originally proposed by \citet{schapireBoostingMarginNew1998}; the boundedness of
these features leads to sub-Gaussian concentration around their mean, similarly
to in \Cref{svm_bound}.

\section{PARTIAL DERANDOMISATION}\label{section:partial-agg}

Bounds of a similar form to \Cref{svm_bound} can also be used in another
interesting situation: where before linear prediction we apply a feature map, as
commonly done in the SVM.\@ If \(\phi \in \Phi \subset \{f: \mathcal{X} \to \mathcal{X}^{\dagger}\}\)
(so that \(\mathcal{X}^{\dagger}\) is a Hilbert space and \(\mathcal{X}\) an arbitrary
set) is the map, our predictor is of the form \(\langle w, \phi(x)\rangle\).
\Cref{svm_bound} then applies with only the modification that \(R\) is a bound
on \(\|\phi(x)\|_2\) instead of \(\|x\|_2\).

In certain cases we may wish to \emph{learn} these (perhaps randomised) features
in parallel with \(w\). In this case the usual PAC-Bayesian analysis would
generally fail without making both \(w\) and the map \(\phi\) random. The
generality of coupling and approximations as outlined in
\Cref{section:derandomising} here comes to the fore; we can ``partially
derandomise'' or derandomise \(w\) while \(\phi\) is still randomised.

More formally, let \(Q^{\Phi} \in \mathcal{M}_1(\Phi)\) be a probability measure
on feature maps so that the posterior \(Q\) is a distribution on functions of
the form \(f(\phi(x))\) for \(\phi \sim Q^{\Phi}\) and deterministic
\(f: \mathcal{X}^{\dagger} \to \mathcal{Y}\). The approximating \(P\)
distribution can then take the form \(g(\phi'(x))\) for \(g \sim P^{g}\) and
\(\phi' \sim Q^{\Phi}\), the same random feature map. Provided the \(P^g\) and
\(Q^{\Phi}\) distributions are independent, the KL divergence from prior \(P_0\)
separates into terms like
\(\KL{P, P_0} = \KL{P^g, P_0^g} + \KL{Q^{\Phi}, P_0^{\Phi}}\). Using this fact
and that such mappings do not affect the sub-Gaussianity of our predictors, we
obtain the following results, analagous to \Cref{svm_bound}, but applicable
under learned and potentially randomised feature maps.

\begin{lemma}\label{lemma:coupling-partial-randomisation}
There is a \(1\)-sub-Gaussian coupling between functions defined by
\(h(x) = \langle w, \phi(x)\rangle\) and \(h'(x) = \langle w + g, \phi'(x)\rangle\) where
\(g \sim \mathcal{N}(\vec{0}, \bm{I})\) and \(\phi, \phi' \sim Q^{\Phi}\) independent of \(g\),
provided \(\|\phi\|_2 \le 1\) almost surely \([Q^{\Phi}]\).
\end{lemma}

\begin{proof}
We use a coupling \(\pi\) such that \(\phi = \phi'\), so that
\begin{align*}
  \E_{(h, h') \sim \pi} &\exp(t (h(x) - h'(x))) \\
  &= \E_{g} \E_{\phi \sim Q^{\Phi}} \exp(t (\langle w, \phi\rangle - \langle w + g, \phi\rangle ))  \\
&= \E_{\phi \sim Q^{\Phi}} \exp(t^2 \|\phi\|_2^2 / 2)
\le \exp(t^2 / 2)
\end{align*}
where we use the moment generating function of a standard
multivariate Gaussian.
\end{proof}

\begin{theorem}
In the binary classification setting, let \(\Phi\) be a space of bounded functions
\(\phi: \mathcal{X} \to \mathcal{X}^{\dagger}\) where \(\mathcal{X}^{\dagger}\) is a Hilbert
space with \(\|\phi\|_2 \le 1\) everywhere. For any prior
\(P_0^{\Phi} \in \mathcal{M}_1(\Phi)\) and \(\delta \in (0, 1)\), with probability
\(\ge 1 - \delta\) over \(S \sim D^m\), for all prediction distributions \(Q\) of the
form \(f(x) = \langle w, \phi(x)\rangle\) with \(\|w\|_2 \le 1, \phi \sim Q^{\Phi} \in \mathcal{M}_1(\Phi)\),
\[ L(Q) \le \hat{L}_{\gamma}(Q) + \sqrt{\frac{\hat{L}_{\gamma}(F_{\vec{w}}) \cdot \Delta}{m}} + \frac{\Delta + \sqrt{\Delta} + 2}{m} \]
where
\(\Delta := 2\log(2/\delta) + 9(R/\gamma)^2 \log m + 2 \KL{Q^{\Phi}, P_0^{\Phi}}.\)
\end{theorem}

\begin{proof}[(Sketch of proof)]
Use \Cref{lemma:coupling-partial-randomisation} in the proof of the second
part of \Cref{svm_bound} to obtain 1-sub-Gaussianity, adding the extra KL
contribution from the feature map.
\end{proof}

Such a bound (with \(L_1\) norm restrictions) could be used to derandomise the
final layer of neural networks with a bounded (e.g. \(\tanh\)) activation
functions on the penultimate layer and randomised weights on the rest of the
structure. In the next section we will take this approach further and
derandomise the final \emph{two} layers through margins, which can be
straightforwardly used to obtain a bound on one-hidden-layer networks. In
conjunction with the above ideas it yields bounds for deep stochastic networks
with the final two layers derandomised.

\section{AVERAGING ONE-HIDDEN-\allowbreak{}LAYER NETWORKS}
\label{section:one-hidden}

In this section we prove generalisation bounds for a one-hidden-layer neural
network (possibly with a randomised input feature map) with a slightly unusual
\(\erf\) activation function that looks much like a \(\tanh\) or other
sigmoidal-type function as more commonly used. This is inspired by the work of
\citet{germainPACBayesianLearningLinear2009} and \citet{NIPS2019_8911}, which
consider averaging over the predictions of functions like
\(f_{w}: \Re^{d} \to \Re, x \mapsto \sign(w \cdot x)\), where
\(w \sim \mathcal{N}(u, I)\), giving ``aggregated'' prediction functions of the
form
\begin{equation}
\label{eq:germainagg}
F(x) = \mathbb{E}_{w \sim \mathcal{N}(u, I)} \sign(w \cdot x) = \erf(u \cdot x / \sqrt{2} \|x\|_{2}).
\end{equation}

With a clever choice of weight distribution, we can combine the sub-Gaussian
concentration of the bounded sign function with its tractable average to get
bounds for one-hidden layer networks.

\begin{definition}
\emph{Single Hidden Erf Layer (SHEL) Network}: Given
\(V \in \Re^{c \times K}\) and \(U \in \Re^{K \times d}\), this is the neural
network \(F: d \to c\) defined by
\begin{equation}\label{def:erfnet}
F_{U, V}(x) = V \erf\left(\frac{U x}{\sqrt{2} \|x\|_{2}}\right)
\end{equation}
where the \(\erf\) activation function is applied elementwise. We also
consider the ``binary'' case, where \(V\) is a vector, \(v \in \Re^{K}\).
\end{definition}

The generalisation bound for this depends on a set of prior parameters (or
``random features''), \(U_0\), chosen independently of the training data, for
example the initialisation of the network (this choice has been extensively
discussed in the literature, beginning with \citealp{dziugaite2017computing}).

\begin{theorem}\label{multiclass_erf_theorem}
Fix prior parameters
\(U^{0} \in \Re^{K \times d}\) and
\(\delta \in (0, 1)\). With probability \(\ge 1 - \delta\)
over \(S \sim \mathcal{D}^{m}\), \(L(F_{U, V})\) is upper bounded by
\begin{equation*}
 \hat{L}_{\gamma}(F_{U, V}) + \tilde{O}\left( \frac{\sqrt{K}}{\gamma \sqrt{m}} \left( V_{\infty}\|U - U^0\|_F + \|V\|_F \right)  \right),
\end{equation*}
for any margin \(\gamma > 0\) and any prediction function \(F_{U, V}\) specified
as in \Cref{def:erfnet} with parameters \(U, V\), and
\(V_{\infty} := \max_{ij}|V_{ij}|\). A full (tighter) expression with constants is
given in the proof in \Cref{multiclass_erf_proof}.
\end{theorem}

\textbf{Remark.} At first glance this bound might appear to grow with width,
since although the norm terms are usually seen to be roughly constant under
increasing \(K\), the \(\sqrt{K}\) term is obviously not. However, this is not
necessarily true: the range of the network (and thus maximum margin) is bounded
by \(KV_{\infty}\), so provided the margin per-unit (\(\gamma/K\) for the
\(\gamma\) used in the bound) remains constant, the bound would actually
decrease with \(K\).

To emphasise this, we note that the above bound is unchanged under two simple
transformations, which ensures dimensional consistency (if it were not, we could
perform these operations to obtain a possibly arbitrarily tight bound). (1)
Scale \(V\); the bound and norm term exactly cancel since we can scale
\(\gamma\) by the same amount and obtain the same empirical margin loss. (2)
Double the width of network, with exact copies of weights in the copy: we can
again double \(\gamma\) for a fixed margin loss, while the squared norms also
double.

\textbf{Proof outline.} The central idea underlying the proof is the
construction of a stochastic neural network with \Cref{def:erfnet} as its
average. We replace the normal distribution of \Cref{eq:germainagg} with a
\emph{mixture} of isotropic Gaussians: if the mixture weights are uniform and
their means are given by the columns of \(U\) (notated as the set
\(\{U_{1;\cdot}, \dots, U_{K; \cdot}\}\)), we note that
\begin{multline}
\label{eq:hiddenlayeragg}
\mathbb{E}_{i \sim \operatorname{Unif}(K),\, w \sim \mathcal{N}(U_{i; \cdot}, I)} \sign(w \cdot x)  = \\ \sum_{k=1}^{K} \erf\left(\frac{U_{k; \cdot} \cdot x}{\sqrt{2} \|x\|_2}\right)
\end{multline}
which is directly proportional to one of the output components of the SHEL
network \(F\). To obtain the final layer weights we multiply the \(\sign\) by a
random vector \(\vec{r}\) supported on \(\{+1, -1\}^c\) and re-scale everything
to fit the scale of the SHEL network.

We note that the function \(f(x) = \sign(w \cdot x)\) is also sub-Gaussian (for
any distribution on \(w\)) as it is a bounded random variable for any fixed
\(z\). To obtain control over the constant and thus \(\epsilon\), we average
over a number of copies of the network, an approach inspired by the approach of
\citet{schapireBoostingMarginNew1998} or \citet{langford01boundsfor}, but for a
hidden-layer network. Combination with \Cref{tightened_version}, careful
bounding of the KL divergence of such hierarchical distributions, and a union
bound over margin values completes the proof.

\textbf{Generalisation to bounded functions.}
We note that in the proof of \Cref{multiclass_erf_theorem} we can replace the
\(\sign\) activation functions used in the proxy function distribution by any
bounded activations, for example sigmoid. Indeed, any feature map which is
bounded and independent from the final layer is possible. The caveat is that the
obtained networks have modified activation functions which may not be
analytically tractable.

\textbf{Partial derandomisation.} A more straightforward extension to deep
networks follows through the partial derandomisation framework discussed in
\Cref{section:partial-agg}; the boundedness of the activation then means the
theorem and proof hold with only slight modification. A simple way to do this is
to ``stack'' our SHEL network on top of a ReLU network with Gaussian weights,
adding only a small KL contribution to the bound; this is discussed further in
\Cref{section:experiments}.

This is interesting because it enables empirical comparisons with deeper
networks on more complex datasets without severe overfitting, which we hope can
form a stepping stone between totally-randomised PAC-Bayesian bounds and
non-random margin bounds, while helping in the understanding of one-hidden-layer
network generalisation. This provides a middle ground between a series of works
obtaining bounds for stochastic neural networks such as
\citet{dziugaite2017computing}, and those providing margin bounds for
non-stochastic DNNs, such as (in a PAC-Bayesian context)
\citet{DBLP:conf/iclr/NeyshaburBS18}.

\textbf{Empirical evaluation.} Although the main contribution of this paper is
in the refinement of methods for proving PAC-Bayes margin bounds, in
\Cref{section:experiments} we also make some empirical evaluations of
\Cref{multiclass_erf_theorem}, and a partially derandomised generalisation of
it. Since these bounds were in general vacuous, we adopt the procedures of
\citet{DBLP:conf/iclr/JiangNMKB20} and \citet{DBLP:conf/nips/DziugaiteDNRCWM20}
to compare such bounds; training to a fixed cross-entropy of \(0.3\) and setting
margin loss \(\hat{L}_{\gamma}(F) = 0.2\), we examine changes in the big-O
complexity measure in \Cref{multiclass_erf_theorem} versus generalisation error
under different hyperparameter changes. Our complexity measure is predictive
under training set size changes and somewhat predictive under learning rate
changes, but like most such measures \citep{DBLP:conf/nips/DziugaiteDNRCWM20},
it is not predictive under changes of width, implying the per-unit margin
decreases significantly with width. We interpret this as follows: at
initialisation \(u_i \cdot x \sim d^{-\frac12}\) is small, so if weights stay
near their initialisation (as is usual for wider networks trained by SGD), units
are less saturated and the per-unit margin decreases. This is avoided in lower
dimensions or by scaling up the weight initialisations with \(d\), but as this
is further from the typical SGD training scenario we avoid this.

\textbf{Optimisation of the prior.} We have in the above empirical evaluation
neglected to utilise optimised data-dependent priors \citep[as initiated
by][]{DBLP:conf/nips/AmbroladzePS06,DBLP:journals/jmlr/Parrado-HernandezASS12},
which has been demonstrated to vastly tighten bounds in the case of neural
networks due to the stability of training. These ideas have been heavily used in
recent papers for neural networks \citep[for
example]{DBLP:conf/aistats/DziugaiteHGA021} and were found to significantly
improve the actual bound values in preliminary experiments, in some cases
leading to non-vacuous (although loose) results. As our focus is more on the
theoretical side of providing a method to prove margin bounds, we decided to
focus on the data-independent case for simplicity.

\textbf{Related Work.} Here we mention previous work
\citep{NIPS2019_8911,biggs2021entropy} on PAC-Bayesian neural networks with
\(\erf\) activations, as well as a wide range of results obtaining
generalisation bounds for neural networks, in particular
\citet{neyshabur2018the} which focuses specifically on one-hidden-layer
networks. \citet{DBLP:journals/corr/abs-2002-09956} uses similar methods to ours
by looking at Gaussian perturbations to the weights of a deep ReLU network, but
their bound relies on the strong assumption of bounds on the Hessian and
gradients of the network across weight values, and as formulated is not
evaluable nor decreases with \(m\).

We also highlight an interesting connection to a strand of work
\citep{DBLP:conf/icml/Kristiadi0H20,DBLP:conf/icml/DaxbergerNAAH21} in the
Bayesian neural network literature, where networks involving only some
randomised weights (effectively, partially-derandomised networks) were found to
offer many of the benefits of more general networks while offering considerable
computational saving.

\section{BEYOND TWO LAYERS}\label{section:beyond-two-layers}

Finally, we give a bound for deep feed-forward ReLU networks, similar in form
and proof to that given by \citet{DBLP:conf/iclr/NeyshaburBS18}. Although the
new result shares the same shortcomings (as discussed in, for example,
\citealp{DBLP:conf/nips/DziugaiteDNRCWM20}), we hope our simplified proof and
unifying perspective will help clear the way for future improvements.

The new bound also replaces a factor of \(d\), the number of layers, with one of
\(\sqrt{\log m}\), while the proof is simplified by merely requiring
\(\AV_{\gamma} \in O(m^{-1})\) rather than \(\AV_{\gamma} = 0\) as in the
original. Bounding this term for simple Gaussian weights with the same
perturbation bound as their proof, gives a simple form for KL divergence.
Combination with \Cref{tightened_version} and a cover of different weight
variances and margins completes the proof, given in
\Cref{section:relu-net-proof}.

\begin{theorem}\label{th:relu-net-bound}
  Let \(F: \mathcal{X} \to \Re^c\) on
  \(\mathcal{X} = \{x \in \Re^d : \|x\|_2 \le R \}\) be a fully-connected,
  feed-forward ReLU neural network with \(d\) layers and no more than \(h\)
  units per layer. For fixed \(\delta \in (0, 1), W_{\star} > 0\) and prior
  weight matrices \(\{W_i^0\}_{i=1}^d\), with probability at least
  \(1 - \delta\) for all such networks \(F\) with weight spectral norms
  \(\|W_i\|_2 \le W_{\star}\) for all \(i\), and \(\theta > 0\), \(L(F)\) is
  upper bounded by
  \begin{equation*}
     \hat{L}_{\theta}(F)+O\left( \sqrt{\frac{h r^2 \log(mdh)}{\theta^2 m} \cdot \sum_{i=1}^d \frac{\|W_i - W_i^0\|_F^2}{\|W_i\|_2^2}  + \frac{A}{m}} \right)
  \end{equation*}
  where \(r := R\prod_{i=1}^d \|W_i\|_2\) is an upper bound on the range of
  network, and \(A := \log\frac{1}{\delta} + d\log\log W_{\star}\).
\end{theorem}

\textbf{Remark.} A second difference between \Cref{th:relu-net-bound} and the
bound of \citet{DBLP:conf/iclr/NeyshaburBS18} is the appearance of the prior
matrices (to bring the bound into line with others which often set these to the
initialisation) and the norm bound \(W_{\star}\). This \(W_{\star}\) term arises
from these prior matrices and can be eliminated if the prior matrices are set to
zero, since re-scaling the weights and margins will then not affect the bound
(due to the positive homogeneity of the ReLU,
\(\|W_i\|_2/\theta = \|\tilde{W}_i\|_2/\tilde{\theta}\) and
\(\|W_i\|_F/\|W\|_2= \|\tilde{W}_i\|_F/\|\tilde{W}\|_2\) for re-scaled
\(\tilde{W}_i\) and \(\tilde{\theta}\)).

\section{CONCLUSION}\label{section:conclusion}

In this work we have provided a unified framework for derandomising PAC-Bayes
bounds using margins. In particular this leads to new bounds or greatly
simplified proofs for a variety of settings. It also enables the novel idea
of partial-derandomisation, which provides a halfway house for estimators which
cannot be so easily derandomised.

Specifically: we provided in \Cref{svm_bound} bounds for \(L_2\)-regularised
linear classification which improve upon classical results and match the
state-of-the-art order given in \citet{DBLP:conf/icml/GronlundKL20} while
providing explicit (and small) constants as well as a considerably simplified
proof. In \Cref{section:partial-agg} we extended this result to the novel
situation where we are simultaneously learning a (randomised) feature map. We
then gave further bounds in \Cref{multiclass_erf_theorem} for the novel setting
of single-hidden-erf-layer (SHEL) networks, as well as a bound in
\Cref{th:relu-net-bound} that improves slightly on \citet[Theorem
1]{DBLP:conf/iclr/NeyshaburBS18}. We feel that SHEL networks have much potential
as a setting to explore margin bounds for deterministic neural
networks---matching contemporary practice---through improved concentration
techniques and priors.

Although we recognise that our final results for linear prediction and deep ReLU
networks are relatively small improvements on existing results, we believe our
radically simplified proofs and explicit link of derandomisation to
concentration (a link which has been occasionally used implicitly in proofs in
the literature) are significant and novel contributions to a difficult and
central problem in their own right. We show in \Cref{subsection:covering} how
reducing PAC-Bayes derandomisation to a covering approach leads to a sub-optimal
dependence on the dimension, which is observed in some prior results such as the
prior ReLU bound \citep{DBLP:conf/iclr/NeyshaburBS18}. We believe that by
highlighting this issue we point the way forward to further simplifications and
improvements, and hope the machine learning and statistics community will
leverage these tools going forward.

Similarly, we feel that a major implication of our work is to show that for
non-vacuous neural network margin bounds, we need tighter bounds on the
concentration properties of networks. Networks are observed to be quite robust
to perturbation in practice, far better than the Lipschitz constant-dependent
bounds of our \Cref{th:relu-net-bound} and \citet{neyshabur2018the} would
suggest. Tighter concentration bounds would immediately lead to improved margin
bounds through our framework and would represent a major contribution to
contemporary statistical learning theory.

%% file: appendix.tex

\section{PAC-BAYES BOUNDS}\label{pac-bayes-theorems-reference}

Here we give three different PAC-Bayesian bounds for losses in \([0, 1]\), as
used in the proof of \Cref{tightened_version}. We also define the convenience
function (for \(C > 0, p \in [0, 1]\))
\[ \Phi_{C}(p) = -\frac{1}{C}\log(1 - p + pe^{-C})\]
which has inverse
\[ \Phi_{C}^{-1}(t) = \frac{1-e^{-Ct}}{1-e^{-C}}. \]

\begin{theorem}
  Given data distribution \(\mathcal{D}\) on \(\mathcal{X} \times \mathcal{Y}\),
  prior \(P_{0} \in \mathcal{M}^{1}(\mathcal{H})\), \(C > 0\) and
  \(\delta \in (0, 1)\), the following hold each with probability
  \(\ge 1 - \delta\) over \(S \sim D^{m}\), for all
  \(Q \in \mathcal{M}^{1}(\mathcal{H})\)

  ``small-kl'' (\citet{langford01boundsfor}, with improvement by
  \citet{DBLP:journals/corr/cs-LG-0411099})
  \begin{equation*}
    \smallkl(\hat{L}(Q)  : L(Q)  )
    \le \frac{1}{m} \left( \KL{Q, P_0} + \log\frac{2\sqrt{m}}{\delta} \right)
  \end{equation*}

  ``Catoni'' \citep{catoni2007}
  \begin{equation*}
    L(Q) \le \Phi_{C}^{-1} \left( \hat{L}(Q) + \frac{\KL{Q, P_0} + \log\frac{1}{\delta}}{Cm} \right)
  \end{equation*}
\end{theorem}

For completeness, we also include here Proposition 2.1 from
\citet{germainPACBayesianLearningLinear2009}.

\begin{lemma}
  For any \(0 \le q \le p < 1\),
  \[ \sup_{C > 0} \left[ C\Phi_{C}(p) - Cq \right] = \smallkl(q : p). \]
\end{lemma}

\begin{lemma}\label{lemma:small-kl-lower-bound}
  For all \(\epsilon \in [0, \frac{1}{2}], p \in [0, 1], p > \epsilon\) (with
  the final condition ensuring the left hand side is well-defined),
    \[\smallkl(\epsilon : p - \epsilon) \ge p + 4\epsilon \log \epsilon.\]
\end{lemma}
\begin{proof}
  Note that \(-\log(p-\epsilon) \ge 0\) if \(p \le 1\) and thus
  \(\epsilon\log\frac{\epsilon}{R-\epsilon} \ge \epsilon \log \epsilon\). Using
  the bound \(\log x \le x - 1\) we also have that
  \((1-\epsilon)\log\frac{1-\epsilon}{1+\epsilon-R} \ge R-2\epsilon\). Combining
  these results,
  \(\smallkl(\epsilon : p - \epsilon) \ge p + \epsilon(\log\epsilon - 2)\);
  combination with the bound
  \(\epsilon(\log \epsilon - 2) \ge 4\epsilon\log\epsilon\) in the specified
  range to completes the proof.
\end{proof}

\section{PROOF OF THEOREM \ref{theorem:cover-lower-bound}}\label{section:cover-lower-bound}

First we prove the following lemma.

\begin{lemma}\label{lemma:restricted-measure}
  Let \(\tilde{\mu}\) be a probability distribution supported only on \(A\), then for any other probability distirbution \(\nu\)
  \[\KL{\tilde{\mu}, \nu} \ge -\log \nu(A). \]
\end{lemma}

\begin{proof}
  For the case \(\nu(A) = 0\) or where \(\tilde{\mu}\) is not absolutely
  continuous w.r.t. \(\nu\) the above holds trivially as the KL is infinite.

  Thus assume \(\nu(A) > 0\) and \(\tilde{\mu} \ll \nu\). Define the restriction
  (or conditional distribution) of \(\nu\) to \(A\) as
  \[\tilde{\nu} =
    \begin{cases}
      \nu/\nu(A) & \text{on A} \\
      0 & \text{else}.\\
      \end{cases}\]

  Given the above assumptions, we have
  \[\frac{d\tilde{\mu}}{d\nu} = \frac{d\tilde{\mu}}{d\tilde{\nu}} \frac{1}{\nu(A)}\]
  so from the definition of and non-negativity of KL divergence,
  \[ \KL{\tilde{\mu}, \nu} = \log\frac{1}{\nu(A)} + \KL{\tilde{\mu}, \tilde{\nu}} \ge -\log \nu(A).\]
\end{proof}

\begin{proof}[Proof of \Cref{theorem:cover-lower-bound}]
  Let \(Q_w\) be a deterministic distribution on as-yet-unspecified vector \(w\).

  \(\kappa(Q_w, P_0; \gamma, 0) = \min_{P \in \mathcal{P}}\KL{P, P_0}\) where
  \begin{align*}
    \mathcal{P} &= \{P : \AV(P, Q_w) = 0\} \\
                &= \{P : \sup_{(x, y) \in \mathcal{Z}} \Pr_{u \sim P}(y\langle w - u, x\rangle > \gamma) = 0\} \\
                &= \{P : \forall (x, y) \in \mathcal{Z}, \forall u \in \operatorname{supp}(P), y\langle w - u, x\rangle \le \gamma\} \\
                &= \{P : \forall u \in \operatorname{supp}(P), \|w - u\|_2 \le \gamma\} \\
                &= \{P : \operatorname{supp}(P) \subset \operatorname{Ball}(w, \gamma)\}.
  \end{align*}

  Combining with the \Cref{lemma:restricted-measure} we find that
  \(\kappa(Q_w, P_0; \gamma, 0) \ge -\log P_0[\operatorname{Ball}(w, \gamma)]\).

  Since \(w\) can be chosen in an adversarial manner based on \(P_0\), \(P_0\)
  must not over-weight any such ball. The \(P_0\) which minimises \(\kappa\)
  over all choices of \(Q_w\) is thus uniform over the set of possible balls
  \(\operatorname{Ball}(w, \gamma)\), which is the set
  \(\operatorname{Ball}(0, 1 + \gamma)\) (since \(\|w\|_2 \le 1\)).

  Basic calculation then shows that
  \[\kappa \ge -\log\frac{\operatorname{vol}[\operatorname{Ball}(w, \gamma)]}{\operatorname{vol}[\operatorname{Ball}(0, 1 + \gamma)]} = d \log\frac{1+\gamma}{\gamma} = \Omega(d).\]
\end{proof}

\section{COMPARISON OF THEOREM \ref{svm_bound} TO EXISTING BOUNDS}\label{section:comparison}

Here we discuss how our results in \Cref{svm_bound} compare to existing
results. All of the following will be in the setting of this theorem with
\(R = 1\) (since all bounds only depend on \(R\) through the scaling
\(R/\gamma\)).

\subsection{Hard Margin Case}

The best existing result for this case is the following, from
\citet{bartlett1998generalization} (In theorem 1.6, they show that
\(\operatorname{fat}(\gamma) \le (R/\gamma)^2\), then they use theorem 1.5; here
we give the explicit constants from this theorem):
\[ L \le \frac{1}{m} \left( \frac{256}{\gamma_{\star}^2} \log{\frac{e m \gamma_{\star}^2}{16}} \log{32 m} +  2 \log{\frac{8m}{\delta}}\right)  \in O\left( \frac{1}{m}\left( \gamma_{\star}^{-2}\log^2 m + \log\frac{1}{\delta}\right)\right). \]

From this we see that not only does \Cref{svm_bound} improve in order by
removing a factor of \(\log m\), but also considerably improves the constant
factors.

\citet{DBLP:conf/icml/GronlundKL20} show that there exists
a dataset, and an estimator with \(\|w\|_2 \le 1\), such that:
\[ L \ge \Omega\left( \frac{\log m}{m \gamma_{\star}^2} \right)\] which is matched by
our \Cref{svm_bound} and confirms it cannot be improved in order without
additional assumptions.

\subsection{Soft Margin}

Several somewhat different results appear in this case. Using Rademacher
complexity, Theorem 21 of \citet{bartlett2002rademacher} implies the following
(where the big-O follows by combination with the trivial bound \(L \le 1\)):

\[ L \le \hat{L}_{\gamma} + \frac{4}{\sqrt{m}\gamma} + \left( \frac{8}{\gamma} + 2 \right) \sqrt{\frac{\log{\frac{4}{\delta}}}{m}} = \hat{L}_{\gamma} + O\left( \sqrt{\frac{\gamma^{-2} + \log(1/\delta)}{m}}\right). \]

Based on a more complex bound in \citet{langford2003pac},
\citet{DBLP:conf/colt/McAllester03} gives the bound (for \(m \ge 4\)):
\begin{align*}
L &\le \hat{L}_{\gamma} + \frac{8}{m\gamma^2} + 2\sqrt{2\left(\frac{\hat{L}_{\gamma}}{m\gamma^2} + \frac{4}{m^2\gamma^4} \right)\log{\frac{m\gamma^2}{4}}}  + O\left(\sqrt{\frac{\log m + \log(1/\delta)}{m}}\right) \\
  &= \hat{L}_{\gamma} + O\left(\frac{\log m}{m\gamma^2} + \sqrt{\frac{\log m}{m\gamma^2} \hat{L}_{\gamma}} + \sqrt{\frac{\log m + \log(1/\delta)}{m}} \right).
\end{align*}
The above also leads to a hard-margin formulation which is however weaker than
the \citet{bartlett1998generalization} bound for all but very tiny margins, as
pointed out by \citet{DBLP:conf/icml/GronlundKL20}.

The state-of-the-art, nearly tight bound given by
\citet{DBLP:conf/icml/GronlundKL20} is the following, not given with any
constants,
\[ L \le \hat{L}_{\gamma} + O\left(\frac{\gamma^{-2}\log m + \log(1/\delta)}{m} + \sqrt{\frac{\gamma^{-2}\log m + \log(1/\delta)}{m} \cdot \hat{L}_{\gamma}}\right)\]
which is shown in the same paper to be nearly-tight, in the existential sense
that there exist data distributions for which it cannot be improved.

This matches exactly the bound given in \Cref{svm_bound} in order, but we
emphasise both the simplicity of our proof and that we give constants, making it
actually evaluable in practice.

\section{LINEAR CLASSIFICATION WITH \(L_1/L_{\infty}\) NORMS}\label{section:l-infty-proof}

Here we provide a bound for \(L_1\)-normed linear predictors, that essentially
replicates the results of \citet{DBLP:journals/ai/GaoZ13a} or
\citet{langford01boundsfor}.

\begin{theorem}\label{boosting_bound}
  In the binary classification setting with \(\mathcal{X} \subset \Re^{K}\) such
  that \(\|x\|_{\infty} \le R\), for any \(\delta \in (0, 1)\), \(m \ge 8\), and
  \(\gamma > 0\), with probability \(\ge 1 - \delta\) over \(S \sim D^{m}\), for
  all linear predictors \(F_w(x) = \sum_{k=1}^K w_k x_k \) with
  \(\|w\|_1 \le 1\)
\[ L(F_{\vec{w}}) \le \hat{L}_{\gamma}(F_{\vec{w}}) + \sqrt{\frac{\hat{L}_{\gamma}(F_{\vec{w}}) \cdot \Delta}{m}} + \frac{\Delta + \sqrt{\Delta} + 2}{m}, \]
  where we define \(\Delta := 2\log(2/\delta) + 19(R/\gamma)^2 \log(2K) \log m\).
\end{theorem}

\begin{proof}[Proof of \Cref{boosting_bound}]
  Without loss of generality (since we can always simultaneously re-scale the
  margin and these) we consider \(R = 1\). For simplicity we will
  also assume initially that the weights are non-negative; negative weights can
  later be included through the standard method of doubling the dimension.

  Our prediction function then has the form \(F(x) = \sum_{k=1}^K w_k x_k\). For
  a fixed margin \(\theta > 0\), we approximate this by unweighted averages of
  the form
  \[ f(x) = \frac{1}{T} \sum_{t=1}^T x_{d(t)} \] where the indices
  \(d(t) \sim P\) for some distribution \(P\) over \([K]\). When \(T\) such
  indices are drawn, we denote this distribution over functions by \(P^T\). As
  an average of \(T\) independent bounded variables, \(P^T\) is
  \((1/T)\)-sub-Gaussian with mean \(F\), and thus by
  \Cref{lemma:coupling-partial-randomisation}
  \[ \AV_{\theta}(P^T, \delta(F)) \le e^{T\theta^2 / 8} = \epsilon. \]
  Choosing \(P_0\) as a uniform distribution on \([K]\) and \(P_0^T\) as \(T\) independent copies of this, we see that
  \[ \KL{P^T, P_0^T} = T \KL{P, P_0} = T \left( \log K - H[w] \right) \le T \log K \]
  where \(H[w]\) is the entropy of a categorical variable with (normalised)
  weights \(w\). This expression using \(H[w]\) could be explicitly used (or
  with a non-uniform prior) to improve the bound, as in
  \citet{seeger2001improved}; we will ignore this here and just use the upper
  bound.

  Setting \(T = \ceil*{8 \theta^{-2} \log m}\) in the small-kl formulation of
  \Cref{tightened_version}, we obtain that
  \begin{equation}
    \smallkl(\hat{L}_{\gamma} + m^{-1} : L - m^{-1} ) \le \frac{1}{m} \left(\ceil*{8 \theta^{-2} \log m} \log K + \frac{1}{2}\log m + \log \frac{2}{\delta}\right) \le \frac{\Delta}{2m}
  \end{equation}
  with probability at least \(1 - \delta\).
  \(\Delta := 19 \theta^{-2} \log K \log m + 2\log(2/\delta)\), since
  \(\theta^{-2} \ge 1\) and \(m \ge 2\) for a non-vacuous bound.

  Relaxing using the lower bound \(\smallkl(q : p) \ge (p-q)^2/(2p)\) for
  \(p > q\) as in the proof of \Cref{svm_bound}, we obtain
  \(L \le \hat{L}_{\gamma} + 2m^{-1} + \sqrt{(\hat{L}_{\gamma} + m^{-1}) \cdot \Delta / m} + \Delta/m\).
  To complete the proof, we allow negative weights by doubling the dimensions.
\end{proof}

\section{PROOF OF THEOREM \ref{multiclass_erf_theorem}}\label{multiclass_erf_proof}

We begin by stating the following useful lemma.

\begin{lemma}\label{lemma:entropy_pm1}
  Let \(X \in \mathcal{M}^1(\{+1, -1\})\) be a random variable with \(\E[X] = x\), and
  \begin{equation*}
    h(x) := \KL{X, \operatorname{Uniform}(\{+1, -1\})}
  \end{equation*}
  the KL divergence from a uniform prior. Then
  \begin{equation*}
    h(x) = \frac{1}{2}\left[ (1+x) \log(1+x) + (1-x) \log(1-x) \right] \le x^2 \log 2.
  \end{equation*}
\end{lemma}

\begin{proof}
  The second equation is simply an explicit statement of the KL divergence. It
  is easy to see from convexity that \(h(x) \le x^2\); the improved (and optimal)
  constant of \(\log 2\) requires a more complex argument, as follows.

  Calculation gives the Maclaurin series
  \[
    (1+x) \log(1+x) + (1-x) \log(1-x) = x^2 + \sum_{n=2}^{\infty} \frac{x^{2n}}{n(2n-1)}
  \]
  which has a radius of convergence of \(1\). Therefore
  \[
    h(x) / x^2 = \frac{1}{2} + \frac{1}{2} \sum_{n=2}^{\infty} \frac{x^{2n}}{(n+1)(2n+1)}
  \]
  which is an increasing function on \((0, 1)\) with supremum \(\log 2\). From
  the definition, \(x \in [-1, 1]\). A similar argument applies for \((-1, 0)\)
  and equality is achieved at \(x=0\), so the inequality holds (and is the
  tightest possible).
\end{proof}

\begin{proof}[Proof of \Cref{multiclass_erf_theorem}]
  Let \(P\) be a probability measure on \(\Re^d \times \{+1, -1\}^c\) defined by
  the following hierarchical procedure: draw a mixture component
  \(k \sim \operatorname{Uniform}(K)\); then \(\vec{w} \in \Re^d\) from a
  Gaussian \(\mathcal{N}(U_{k, \cdot}, I)\) (with mean vector as a row of \(U\))
  and for \(i \in [c]\) draw a component of \(\vec{r} \in \{+1, -1\}^c\) such
  that \(\E\vec{r}[i] = V_{ik}/V_{\infty}\). A sample from \(P\) is a tuple
  \((\vec{w}, \vec{r})\).

  \(P^T\) is then defined for \(T \in \mathbb{N}\) as a distribution on
  functions \(f: \Re^d \to \Re^c\), of the following form:
  \begin{equation*}
    f(\vec{x}) = \frac{1}{T} \sum_{t=1}^T \sign(\vec{w}^t \cdot \vec{x}) \vec{r}^t
  \end{equation*}
  for \(T\) independently drawn samples \((\vec{w}^{t}, \vec{r}^{t}) \sim P\).
  It is straightforward to see that
  \(F(\vec{x}) = V_{\infty}K \cdot \E_{P^T}[f(\vec{x})]\) and therefore
  \(L_{\theta'}(F) = L_{\theta}(\mathbb{E}[f])\) where
  \(\theta' = \theta V_{\infty}K\). Further, since \(P^T\) is the average of \(T\)
  independent bounded variables for any fixed \(\vec{x}\), it is
  \((1/T)\)-sub-Gaussian.

  Thus, for any fixed \(T\), \(\theta > 0\) and prior \(P_0^T\), we have by
  \Cref{tightened_version} and \Cref{sub-gaussian-lemma} that

  \begin{equation*}
    \smallkl(\hat{L}_{\theta'}(F) + e^{-T\theta^2/16} : L(F) - e^{-T\theta^2/16} )
    \le \frac{1}{m} \left( \KL{P^T, P_0^T} + \log\frac{2\sqrt{m}}{\delta} \right).
  \end{equation*}

  We now define a prior distribution on individual parameters \(P_0\), and
  functions \(P_0^T\), in a similar way, but with the distribution over each
  component of \(\vec{r}\) uniform on \(\{+1, -1\}\), and the Gaussian mixture
  means as rows of the data-free matrix \(U^0\). Since the samples are
  independently drawn and the distributions \(P, P_0\) over parameters imply those
  over functions, \(\KL{P^T, P_0^T} \le T \KL{P, P_0}\).

  \(P_0\) and \(P\) can be seen as distributions on
  \(([K] \times \Re^d \times \{+1, -1\}^c)\) with the index \(k \in [K]\)
  marginalised out. From the chain rule for conditional entropy and
  \Cref{lemma:entropy_pm1}, (in a slight abuse of notation since \(P_0, P\) are
  not necessarily densities)
  \begin{align*}
    \KL{P(\vec{w}, \vec{r}), P_0(\vec{w}, \vec{r})}
    &\le \KL{P(k, \vec{w}, \vec{r}), P_0(k, \vec{w}, \vec{r})} \\
    &= \KL{P(k), P_0(k)} + \KL{P(\vec{w}, \vec{r} | k), P_0(\vec{w}, \vec{r} | k)} \\
    &= \KL{P(\vec{w}, \vec{r} | k), P_0(\vec{w}, \vec{r} | k)} \\
    &= \KL{P(\vec{w} | k), P_0(\vec{w} | k)} + \KL{P(\vec{r} | k), P_0(\vec{r} | k)} \\
    &= \frac{1}{K} \sum_{k=1}^K \frac{\|U_{k, \cdot} - U^0_{k, \cdot}\|^2_2}{2} + \frac{1}{K} \sum_{k=1}^K \sum_{i=1}^c h(V_{ik}/V_{\infty}) \\
    &\le \frac{\|U - U^0\|_F^2}{2 K} + \frac{\|V\|_F^2}{V_{\infty}^2 K}\log 2.
  \end{align*}

  For any fixed \(\theta = \gamma/(V_{\infty}K) > 0\) and \(m' > 2\), we set
  \(T = \ceil*{16 \theta^{-2} \log m'}\). The following then holds with
  probability at least \(1 - \delta\):

  \begin{align*}
    m \cdot \smallkl\left( \hat{L}_{\theta'}(F) + \frac{1}{m'} : L - \frac{1}{m'} \right)
    &\le \ceil*{16 \theta^{-2} \log m' } \left( \frac{\|U - U^0\|_F^2}{2 K} + \frac{\|V\|_F^2}{V_{\infty}^2 K}\log 2 \right) + \log \frac{2\sqrt{m}}{\delta}. \\
  \end{align*}

  It remains to cover possible values of \(\theta\). Firstly we note that for
  \(\theta \ge 1\) the bound is trivially true by the boundedness of
  \(f(\vec{x})\), and thus we need only consider \(\theta^{-2} > 1\).

  For \(\alpha > 1\) and \(i = 0, 1, \dots\), set \(\theta_i = \alpha^{-i}\) and
  \(\delta_i = \delta/2(i+1)^2\). Applying the union bound over the above
  equation with these parameters we get that with probability at least
  \(1 - (\pi^2/6)\delta \ge 1 - 2\delta\) that the above is true for each pair
  of \(\theta_i\) and \(\delta_i\). We choose the largest \(\theta_i\) such that
  \(\theta_i \le \theta < \theta_{i-1}\), so that
  \(i \le 1 - \log_{a}(\theta)\). Since
  \(\hat{L}_{\theta} \le \hat{L}_{\theta_i}\) is increasing,
  \(1/\theta_i \le a/\theta\), and
  \(\log(1/\delta_i) \le \log(1/\delta) + 2\log(2 + \log_{a}(1/\theta)) = \log(1/\delta) + 2\log(\log(\alpha^2/\theta)/\log(\alpha))\), we
  finally obtain with probability \(1-\delta\)

  \begin{align*}
    m \cdot \smallkl\left( \hat{L}_{\gamma}(F) + \frac{1}{m'} : L - \frac{1}{m'} \right)
    \le &17 \left(\frac{\alpha V_{\infty}K}{\gamma}\right)^2 \left( \frac{\|U - U^0\|_F^2}{2 K} + \frac{\|V\|_F^2}{V_{\infty}^2 K}\log 2 \right) \log m' \\
    &+ \log \frac{4\sqrt{m}}{\delta} + 2\log\left(\frac{\log(\alpha^2 V_{\infty}K/\gamma)}{\log \alpha}\right) \\
  \end{align*}

  for all weight matrices and every \(\gamma > 0\), and fixed
  \(K > 0, \alpha > 1\).

  Relaxing the bound with Pinsker's inequality \(\smallkl(a:b) \ge (a-b)^2\) and
  setting \(m' = m\) and \(\alpha = 2\) completes the proof.
\end{proof}

\section{PROOF OF THEOREM \ref{th:relu-net-bound}}\label{section:relu-net-proof}

Here we give the proof of \Cref{th:relu-net-bound}, beginning with two lemmas
used.

\begin{lemma}[\citealp{DBLP:conf/iclr/NeyshaburBS18}; Lemma 2, Perturbation Bound.]\label{neyshabur_lemma_2}
  In the setting of \Cref{th:relu-net-bound}, for any layer weights \(W_i\),
  \(x \in \mathcal{X}\) and weight perturbations \(U_i\) such that
  \(\|U_i\|_2 \le d^{-1} \|W_i\|_2\),
  \[
    \| f(x) - F(x) \|_2 \le e R \left( \prod_{i=1}^d \|W_i\|_2\right) \sum_{i=1}^d \frac{\|U_i\|_2}{\|W_i\|_2}
  \]
  where \(F\) is the unperturbed and \(f\) the perturbed network (with weights
  \(W_i\) and \(W_i + U_i\) respectively).
\end{lemma}

\begin{lemma}\label{lemma:av-deep-relu}
  Let \(Q = \delta(F)\) for such a feed-forward ReLU network with weights
  \(W_i\), and \(P\) be the same network with Gaussian weights, with per-layer
  means \(W_i\) and variances \(\sigma_i^2\). Then for all
  \(0 < \theta < 2\sup_{x \in \mathcal{X}, y \in [K]} |F(x)[y]|\),
  \begin{equation*}
    \AV_{\theta}(P, Q) \le 2h \sum_{i=1}^d \exp\left(-\frac{1}{32h}  \left( \frac{\theta \|W_i\|_2}{\sigma_i e R (\prod_i \|W_i\|_2)} \right)^2 \right).
  \end{equation*}
\end{lemma}

\begin{proof}
  From \Cref{neyshabur_lemma_2}, we see immediately that if for all \(i\), the
  perturbations have \(\|U_i\|_2 \le c \theta \|W_i\|_2\) for
  \(c^{-1} = 4 edR \left(\prod_i \|W_i\|_2 \right)\), then
  \(\| f(x) - g(x) \|_2 \le \theta / 4\). The perturbation condition of
  \Cref{neyshabur_lemma_2} is satisfied if \(\theta < 2 eR \prod_i \|W_i\|_2\),
  which is true for any \(\theta\) in the range of the function margins (as in
  the lemma assumption, since \(R\prod_i \|W_i\|_2\) is an upper bound on the
  range). If the perturbations are randomised, we see that (letting \(y' \ne y\)
  achieve the maximum margin)

  \begin{align*}
    \AV_{\theta}(P, Q) &\le \Pr\{|M(f, z) - M(g, z)| > \theta / 2 \} \\
      &\le \Pr\{|f(x)[y] - f(x)[y'] - g(x)[y] + g(x)[y']| > \theta / 2 \} \\
      &\le \Pr\{2\|f(x) - g(x)\|_{\infty} > \theta / 2 \} \\
      &\le \Pr\{\|f(x) - g(x)\|_2 > \theta / 4 \} \\
      &\le \Pr\{\exists i : \|U_i\|_2 > c\theta \|W_i\|_2\} \\
      &\le \sum_{i=1}^d \Pr\{\|U_i\|_2 > c\theta \|W_i\|_2\}.
  \end{align*}

  We set the weights of \(g\) to be Gaussian with diagonal covariance, and
  per-layer variances of \(\sigma_i^2\). To complete the proof we use a result
  of \citet{DBLP:journals/focm/Tropp12} for Gaussian random matrices, that
  \[ \Pr\{\|U_i\|_2 > t\} \le 2 h e^{-t^2 / 2h\sigma_i^2}.\]
\end{proof}

\begin{proof}[Proof of \Cref{th:relu-net-bound}]
  We choose \(P\) and \(P^0\) to have Gaussian weight matrices with means
  \(W_i\) and \(W_i^0\), and identical per-layer variances \(\sigma_i\). From
  \Cref{lemma:av-deep-relu} we have for any fixed \(\theta\) and set of
  \(\sigma_i\), and for all \(F\) with weights \(W_i\), such that the inverse
  variances
  \begin{equation}\label{eq:bounds-inverse-variances}
    \sigma_i^{-2} \ge 32h \left( \frac{e R (\prod_i \|W_i\|_2)}{\theta \|W_i\|_2} \right)^2 \log(mhd)
  \end{equation}
  we have \(\AV_{\theta} \le 2/m\). Therefore from \Cref{tightened_version} and
  Pinsker's inequality we have the generalisation bound (for the weight matrices
  and \(\theta\) satisfying the condition on the set of \(\sigma_i\))
  \begin{equation*}
    L(F) \le \hat{L}_{\theta}(F) + \frac{2}{m} + \sqrt{\frac{1}{2m}\left(\sum_{i=1}^d \frac{\|W_i - W_i^0\|_F^2}{\sigma_i^2} +\log\frac{2\sqrt{m}}{\delta} \right)}.
  \end{equation*}

  We complete the proof by constructing covers for \(\sigma_i\) and \(\theta\).
  We only need to consider \(\theta < R \prod_i \|W_i\|_2 =: C_\theta\) (an
  upper bound on the range of the function) as otherwise the
  \(\hat{L}_{\theta}\) term is \(1\) and the bound is vacuous. Since
  \(\|W_i\|_2 \le W_{\star}\) for all \(i\) we have that
  \(\sigma_i^{-2} \ge 32e^2 h \|W_i\|_2^{-2} \ge 32e^2 h W_{\star}^{-2}\) and
  \(\sigma_i \le 15 h^{-1/2} W_{\star}^2 =: C_\sigma\).

  For \(t = 0, 1, 2, \dots\) choose margins \(\theta^{(t)} = C_{\theta}/2^t\)
  and let the bound for this margin hold with probability
  \(\delta^{(t)} = \delta/(t+1)^2\), so that taking a union bound the above
  holds simultaneously for every \(\theta^{(t)}\) with probability at least
  \(1 - \pi^2 \delta / 6 \ge 1 - 2\delta\). To find a bound holding
  simultaneously for all \(\theta\), we choose the \(t\) such that
  \(\theta^{(t)} \le \theta < \theta^{(t-1)}\), and then replace this term with
  \(\theta\) by using the facts that
  \(\hat{L}_{\theta^{(t)}} \le \hat{L}_{\theta}\),
  \(1/\theta^{(t)} \le 2/\theta\), and
  \(\log(1/\delta^{(t)}) \le \log(1/\delta) + 2\log\log_2(4C_{\theta}/\theta)\).

  Repeating this same covering process for every choice of \(\sigma_i\), we
  obtain with probability at least \(1 - 2\delta\) simultaneously for all
  \(\theta, \sigma_i\) (and thus also for the tightest \(\sigma_i\) satisfying
  \Cref{eq:bounds-inverse-variances}) that \(L(F) - \hat{L}_{\theta}(F)\) is
  upper bounded by
  \begin{align*}
    & \frac{2}{m} + \sqrt{\frac{1}{2m}\left(4\sum_{i=1}^d \frac{\|W_i - W_i^0\|_F^2}{\sigma_i^2} + \log\frac{2(d+1)\sqrt{m}}{\delta} + 2\log\log_2(4C_{\theta}/\theta) + \sum_{i=1}^d 2\log\log_2(4C_{\sigma}/\sigma_i) \right) } \\
    &\in O\left( \sqrt{\frac{h R^2 \left(\prod_{i=1}^d \|W_i\|_2^2\right) \log(mdh)}{\theta^2 m} \cdot \sum_{i=1}^d \frac{\|W_i - W_i^0\|_F^2}{\|W_i\|_2^2}  + \frac{\log\frac{1}{\delta} + d\log\log W_{\star}}{m}} \right)
  \end{align*}
\end{proof}

\section{EMPIRICAL EVALUATION OF \Cref{multiclass_erf_theorem}}\label{section:experiments}

\subsection{Experimental setup}\label{section:experimental-setup}

All experiments were performed using the Tensorflow 2 library
\citep{tensorflow2015-whitepaper} in Python, on a single workstation with a
Nvidia RTX 2080 Ti GPU. Code for the results is licensed under an MIT license
and available in the supplementary material.


We train SHEL networks and a partially-aggregated variation thereof under
different hyperparameter configurations. We use this to compare changes in the
generalisation error (the difference between test and train misclassification
errors) with the complexity term from \Cref{multiclass_erf_theorem} given by
\begin{equation}\label{equation:complexity-term}
  \frac{\sqrt{K}}{\gamma \sqrt{m}} \left( V_{\infty}\|U - U^0\|_F + \|V\|_F \right).
\end{equation}

Following previous empirical evaluations of such complexity terms, we train to a
fixed value of cross-entropy; see \citet{DBLP:conf/iclr/JiangNMKB20} for further
discussion. The margin \(\gamma\) is set as that giving a fixed
\(L_{\gamma}(F) = 0.2\), or \(E_{f \sim Q}L_{\gamma}(f) = 0.2\) for the
partially aggregated version.

For the partially-aggregated version, we include a feature map of three
additional dense ReLU layers with Gaussian weight matrices with independent
components, means \(\{W_i\}_{i=1}^3\) and variances of \(\sigma\). Again using
the initialisation as a prior, this adds a term of
\[\sqrt{\sum_{i=1}^3 \|W_i - W_i^0\|_F^2 / 4m\sigma^2}\]
to the right hand side
of the bound. To enable comparison, we set \(\sigma\) to make this term constant
and equal to a half when calculating \(E_{f \sim Q} L_{\gamma}(f)\). This is done
during the evaluation phase, and training is performed on the non-stochastic
version (weights as means) as in \citet{dziugaite2017computing}.

These experiments aim to evaluate the predictive ability of this complexity
measure under changes of procedures. To this end we provide plots of the
generalisation, \(G(\omega)\), and complexity measure, \(C(\omega)\), for
trained parameters \(\omega\) versus some change in hyperparameter value.

We also provide evaluations using the sign-error, a measure of predictive power
defined in \citet{DBLP:conf/nips/DziugaiteDNRCWM20} as
\begin{equation*}
  \frac{1}{2} \E_{\omega, \omega'} [1 - \sign(C(\omega') - C(\omega)) \cdot \sign(G(\omega') - G(\omega))]
\end{equation*}
where \(\omega\) and \(\omega'\) are parameters obtained through training with
one changed hyperparameter between them. The maximum over such pairs of
hyperparameter settings is a measure of the robustness of predictions made about
the generalisation based on the complexity measure; if this value is low, the
complexity measure makes robust predictions. We provide this maximum, the
median, and the mean of the above \citep[as in][]{DBLP:conf/iclr/JiangNMKB20} for
different setups and allowing different hyperparameters to vary.

\subsection{SHEL Network}

\begin{figure}[p]
  \centering
  \includegraphics[width=\textwidth]{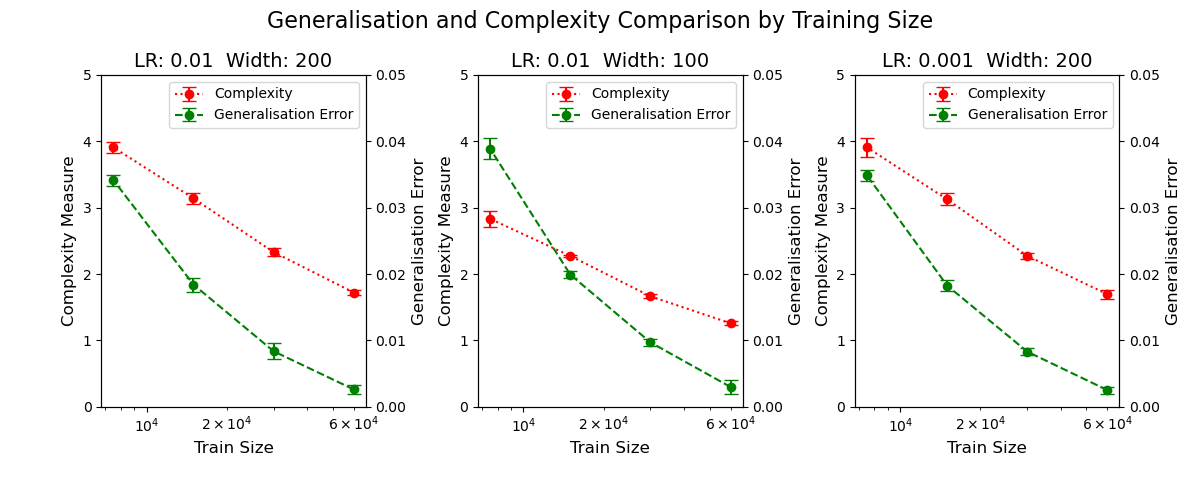}
  \caption{Changes in complexity measure and generalisation error versus training set size under fixed other hyperparameters, for a SHEL network trained on MNIST.}
  \label{fig:tsize}
\end{figure}

\begin{figure}[p]
  \centering
  \includegraphics[width=\textwidth]{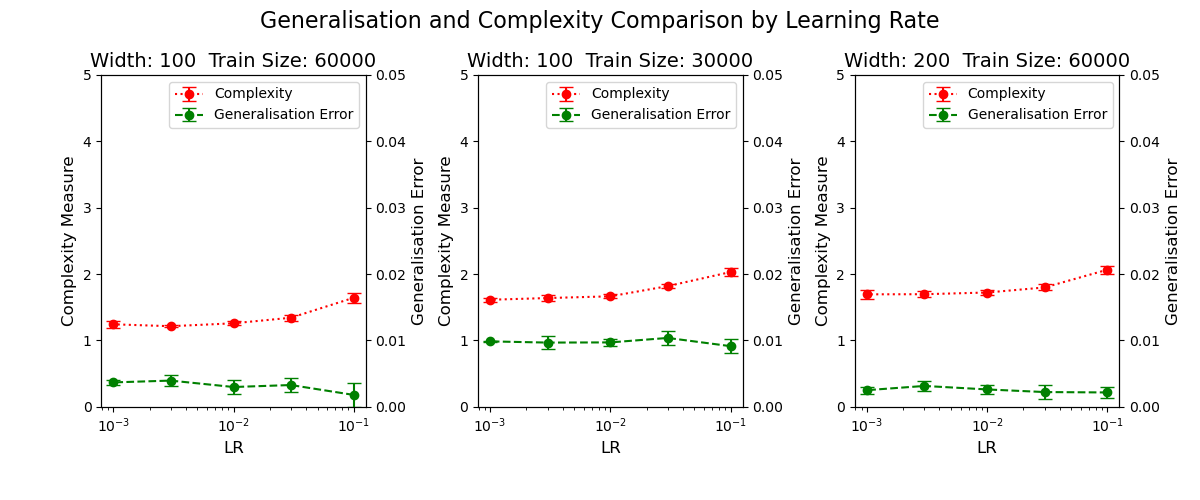}
  \caption{Changes in complexity measure and generalisation error versus learning rate under fixed other hyperparameters, for a SHEL network trained on MNIST.}
  \label{fig:lr}
\end{figure}

\begin{figure}[p]
  \centering
  \includegraphics[width=\textwidth]{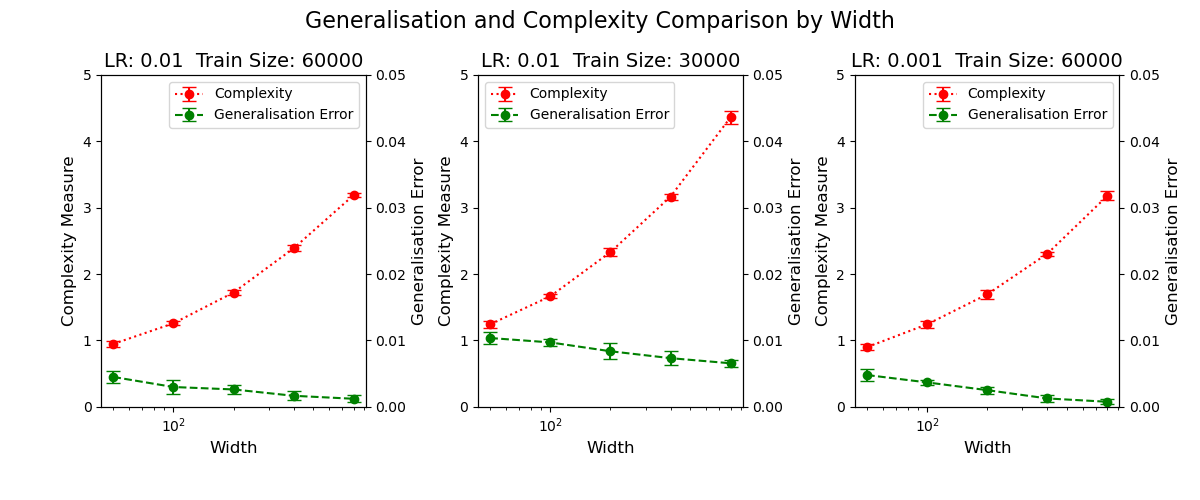}
  \caption{Changes in complexity measure and generalisation error versus width under fixed other hyperparameters, for a SHEL network trained on MNIST.}
  \label{fig:width}
\end{figure}

On the MNIST \citep{lecun2010mnist} dataset, we examine the following
hyperparameter settings, finding through the sign error (\Cref{table:erf-mnist})
that predictions under changes of training size are quite robust, while those
under changes of learning rate or width are poor. We additionally provide plots
(\Cref{fig:width,fig:tsize,fig:lr}) for some selected hyperparameter values to
verify the above. This poor prediction under such changes is unfortunately a
feature of many such complexity measures
\citep{DBLP:conf/nips/DziugaiteDNRCWM20}.

\begin{itemize}
  \item Learning rate \(\in \{10^{-3}, 3 \times 10^{-3}, 10^{-2}, 3 \times 10^{-2}, 10^{-1}\}\).
  \item Train set sizes \(\in \{60\,000, 30\,000, 15\,000, 7\,500\}\).
  \item Width \(\in \{50, 100, 200, 400, 800\}\).
  \item Batch size \(= 200\).
  \item Learning algorithm SGD with momentum parameter \(= 0.9\).
\end{itemize}

\begin{table}[h]
  \centering
  \begin{tabular}{ c c c c }
  \toprule
  Variable Hyperparameter & Max SE & Median SE & Mean SE \\
  \midrule
  Learning Rate & 1.0 & 0.60 & 0.56 \\
  Width & 1.0 & 1.0 & 0.90 \\
  Train Size & 0.2 & 0.0 & 0.00 \\
  All & 1.0 & 0.60 & 0.53 \\
  \bottomrule
  \end{tabular}
  \vspace{0.5em}
  \caption{Statistics of the sign error, SE, under different varying hyperparameters for a SHEL network trained on MNIST.}
  \label{table:erf-mnist}
\end{table}

\subsection{Partially-Derandomised SHEL}

\begin{figure}[p]
  \centering
  \includegraphics[width=\textwidth]{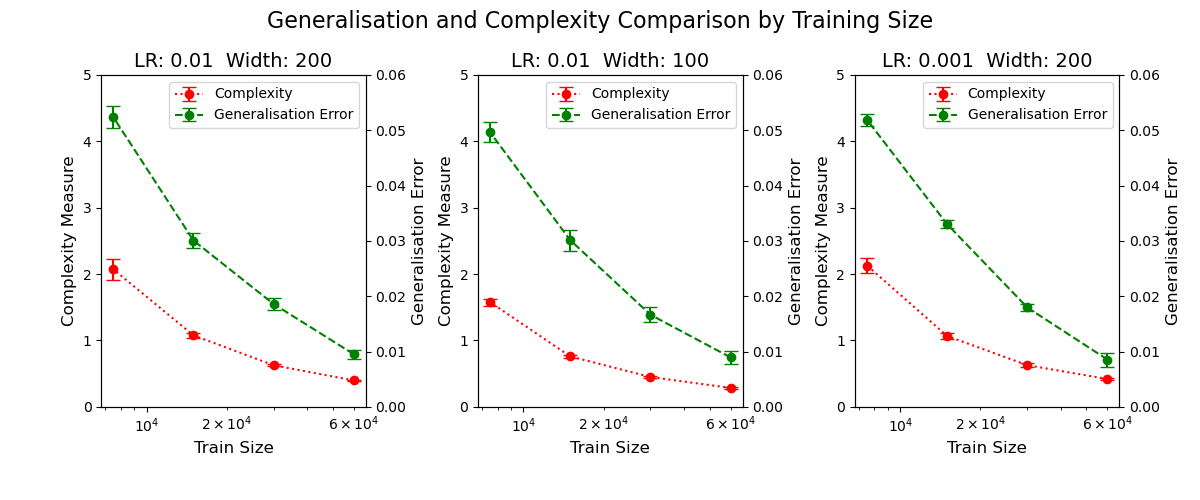}
  \caption{Changes in complexity measure and generalisation error versus training set size under fixed other hyperparameters, for a partially-derandomised SHEL network trained on MNIST.}
  \label{fig:tsize-p}
\end{figure}

\begin{figure}[p]
  \centering
  \includegraphics[width=\textwidth]{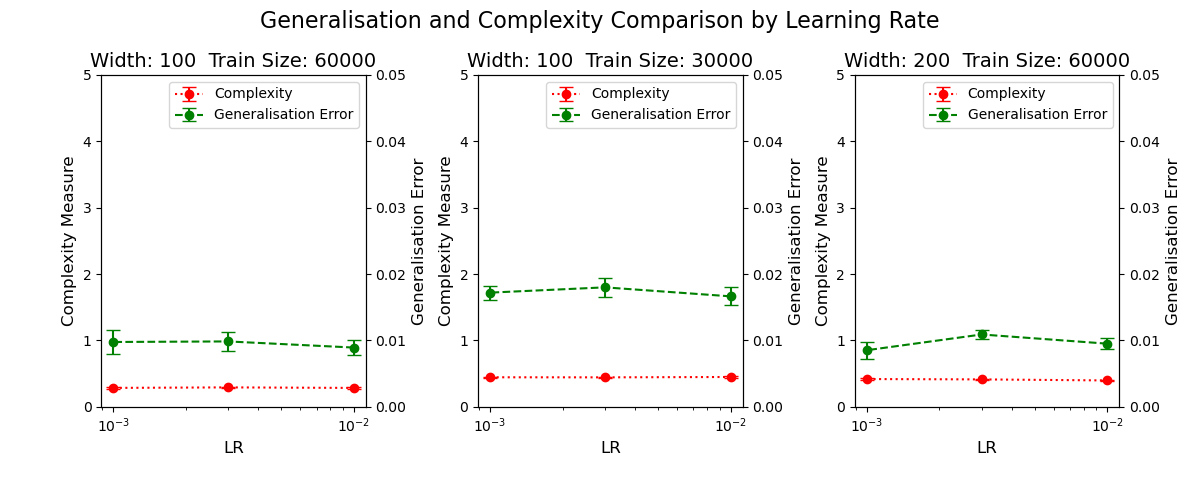}
  \caption{Changes in complexity measure and generalisation error versus learning rate under fixed other hyperparameters, for a partially-derandomised SHEL network trained on MNIST.}
  \label{fig:lr-p}
\end{figure}

\begin{figure}[p]
  \centering
  \includegraphics[width=\textwidth]{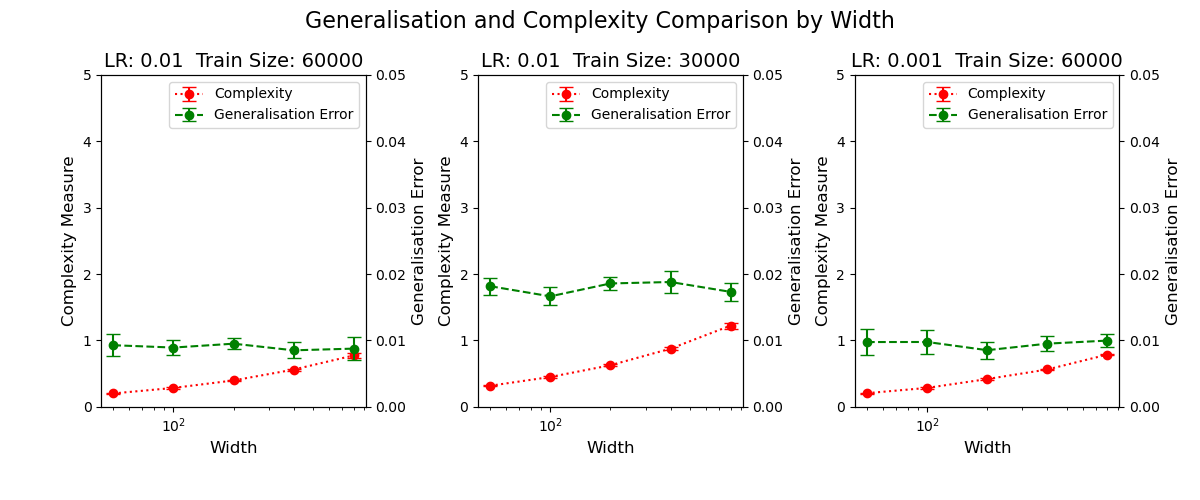}
  \caption{Changes in complexity measure and generalisation error versus width under fixed other hyperparameters, for a partially-derandomised SHEL network trained on MNIST.}
  \label{fig:width-p}
\end{figure}

Again on the MNIST dataset, we evaluate the partially-derandomised version of the
above under the same hyperparameter values, excluding learning rates of \(0.1\)
and \(0.03\) which sometimes led to numerical instability.
\Cref{fig:width-p,fig:lr-p,fig:tsize-p} provide sample results and the
sign-error results are reported in \Cref{table:partial-mnist}.

These sign error results show that predictions under changes of training size
are completely robust, while those under changes of learning rate or width are
still poor. The predictions for width are somewhat improved, though we note that
our estimate of this quantity may be somewhat noisy as the generalisation error
appears largely independent of width.

\begin{table}[h]
  \centering
  \begin{tabular}{ c c c c }
  \toprule
  Variable Hyperparameter & Max SE & Median SE & Mean SE \\
  \midrule
  Learning Rate & 1.0 & 0.60 & 0.49 \\
  Width & 1.0 & 0.40 & 0.46 \\
  Train Size & 0.0 & 0.0 & 0.0 \\
  All & 1.0 & 0.20 & 0.31 \\
  \bottomrule
  \end{tabular}
  \vspace{0.5em}
  \caption{Statistics of the sign error, SE, under different varying hyperparameters for a partially-derandomised SHEL network trained on MNIST.}
  \label{table:partial-mnist}
\end{table}

%% file: main.bbl
\begin{thebibliography}{43}
\providecommand{\natexlab}[1]{#1}
\providecommand{\url}[1]{\texttt{#1}}
\expandafter\ifx\csname urlstyle\endcsname\relax
  \providecommand{\doi}[1]{doi: #1}\else
  \providecommand{\doi}{doi: \begingroup \urlstyle{rm}\Url}\fi

\bibitem[Abadi et~al.(2015)Abadi, Agarwal, Barham, Brevdo, Chen, Citro,
  Corrado, Davis, Dean, Devin, Ghemawat, Goodfellow, Harp, Irving, Isard, Jia,
  Jozefowicz, Kaiser, Kudlur, Levenberg, Man\'{e}, Monga, Moore, Murray, Olah,
  Schuster, Shlens, Steiner, Sutskever, Talwar, Tucker, Vanhoucke, Vasudevan,
  Vi\'{e}gas, Vinyals, Warden, Wattenberg, Wicke, Yu, and
  Zheng]{tensorflow2015-whitepaper}
Mart\'{\i}n Abadi, Ashish Agarwal, Paul Barham, Eugene Brevdo, Zhifeng Chen,
  Craig Citro, Greg~S. Corrado, Andy Davis, Jeffrey Dean, Matthieu Devin,
  Sanjay Ghemawat, Ian Goodfellow, Andrew Harp, Geoffrey Irving, Michael Isard,
  Yangqing Jia, Rafal Jozefowicz, Lukasz Kaiser, Manjunath Kudlur, Josh
  Levenberg, Dandelion Man\'{e}, Rajat Monga, Sherry Moore, Derek Murray, Chris
  Olah, Mike Schuster, Jonathon Shlens, Benoit Steiner, Ilya Sutskever, Kunal
  Talwar, Paul Tucker, Vincent Vanhoucke, Vijay Vasudevan, Fernanda Vi\'{e}gas,
  Oriol Vinyals, Pete Warden, Martin Wattenberg, Martin Wicke, Yuan Yu, and
  Xiaoqiang Zheng.
\newblock {TensorFlow}: Large-scale machine learning on heterogeneous systems,
  2015.
\newblock URL \url{https://www.tensorflow.org/}.
\newblock Software available from tensorflow.org.

\bibitem[Alquier and Biau(2013)]{AB2013}
Pierre Alquier and G\'erard Biau.
\newblock Sparse single-index model.
\newblock \emph{Journal of Machine Learning Research}, 14:\penalty0 243--280,
  2013.

\bibitem[Ambroladze et~al.(2006)Ambroladze, Parrado{-}Hern{\'{a}}ndez, and
  Shawe{-}Taylor]{DBLP:conf/nips/AmbroladzePS06}
Amiran Ambroladze, Emilio Parrado{-}Hern{\'{a}}ndez, and John Shawe{-}Taylor.
\newblock Tighter pac-bayes bounds.
\newblock In Bernhard Sch{\"{o}}lkopf, John~C. Platt, and Thomas Hofmann,
  editors, \emph{Advances in Neural Information Processing Systems 19,
  Proceedings of the Twentieth Annual Conference on Neural Information
  Processing Systems, Vancouver, British Columbia, Canada, December 4-7, 2006},
  pages 9--16. {MIT} Press, 2006.
\newblock URL
  \url{https://proceedings.neurips.cc/paper/2006/hash/3f5ee243547dee91fbd053c1c4a845aa-Abstract.html}.

\bibitem[Banerjee et~al.(2020)Banerjee, Chen, and
  Zhou]{DBLP:journals/corr/abs-2002-09956}
Arindam Banerjee, Tiancong Chen, and Yingxue Zhou.
\newblock De-randomized pac-bayes margin bounds: Applications to non-convex and
  non-smooth predictors.
\newblock \emph{CoRR}, abs/2002.09956, 2020.
\newblock URL \url{https://arxiv.org/abs/2002.09956}.

\bibitem[Bartlett and Shawe-Taylor(1998)]{bartlett1998generalization}
Peter Bartlett and John Shawe-Taylor.
\newblock Generalization performance of support vector machines and other
  pattern classifiers.
\newblock In Bernard Sch{\"o}lkopf, Christopher J~C Burges, and Alexander~J
  Smola, editors, \emph{Advances in Kernel Methods - Support Vector Learning}.
  MIT Press, Cambridge, USA, 1998.

\bibitem[Bartlett and Mendelson(2002)]{bartlett2002rademacher}
Peter~L Bartlett and Shahar Mendelson.
\newblock Rademacher and {{Gaussian}} complexities: {{Risk}} bounds and
  structural results.
\newblock \emph{Journal of Machine Learning Research}, 3:\penalty0 463--482,
  2002.

\bibitem[Biggs and Guedj(2021)]{biggs2021entropy}
Felix Biggs and Benjamin Guedj.
\newblock Differentiable pac–bayes objectives with partially aggregated
  neural networks.
\newblock \emph{Entropy}, 23\penalty0 (10), 2021.
\newblock ISSN 1099-4300.
\newblock \doi{10.3390/e23101280}.
\newblock URL \url{https://www.mdpi.com/1099-4300/23/10/1280}.

\bibitem[Boucheron et~al.(2013)Boucheron, Lugosi, and
  Massart]{DBLP:books/daglib/0035704}
St{\'{e}}phane Boucheron, G{\'{a}}bor Lugosi, and Pascal Massart.
\newblock \emph{Concentration Inequalities - {A} Nonasymptotic Theory of
  Independence}.
\newblock Oxford University Press, 2013.
\newblock ISBN 978-0-19-953525-5.
\newblock \doi{10.1093/acprof:oso/9780199535255.001.0001}.
\newblock URL \url{https://doi.org/10.1093/acprof:oso/9780199535255.001.0001}.

\bibitem[Catoni(2007)]{catoni2007}
Olivier Catoni.
\newblock \emph{{PAC-Bayesian Supervised Classification: The Thermodynamics of
  Statistical Learning}}.
\newblock Institute of Mathematical Statistics lecture notes-monograph series.
  Institute of Mathematical Statistics, 2007.
\newblock ISBN 9780940600720.
\newblock URL \url{https://books.google.fr/books?id=acnaAAAAMAAJ}.

\bibitem[Cortes and Vapnik(1995)]{DBLP:journals/ml/CortesV95}
Corinna Cortes and Vladimir Vapnik.
\newblock Support-vector networks.
\newblock \emph{Mach. Learn.}, 20\penalty0 (3):\penalty0 273--297, 1995.
\newblock \doi{10.1007/BF00994018}.
\newblock URL \url{https://doi.org/10.1007/BF00994018}.

\bibitem[Daxberger et~al.(2021)Daxberger, Nalisnick, Allingham, Antor{\'{a}}n,
  and Hern{\'{a}}ndez{-}Lobato]{DBLP:conf/icml/DaxbergerNAAH21}
Erik~A. Daxberger, Eric~T. Nalisnick, James~Urquhart Allingham, Javier
  Antor{\'{a}}n, and Jos{\'{e}}~Miguel Hern{\'{a}}ndez{-}Lobato.
\newblock Bayesian deep learning via subnetwork inference.
\newblock In Marina Meila and Tong Zhang, editors, \emph{Proceedings of the
  38th International Conference on Machine Learning, {ICML} 2021, 18-24 July
  2021, Virtual Event}, volume 139 of \emph{Proceedings of Machine Learning
  Research}, pages 2510--2521. {PMLR}, 2021.
\newblock URL \url{http://proceedings.mlr.press/v139/daxberger21a.html}.

\bibitem[Dziugaite and Roy(2017)]{dziugaite2017computing}
Gintare~Karolina Dziugaite and Daniel~M Roy.
\newblock Computing nonvacuous generalization bounds for deep (stochastic)
  neural networks with many more parameters than training data.
\newblock \emph{Conference on Uncertainty in Artificial Intelligence 33.},
  2017.

\bibitem[Dziugaite and Roy(2018)]{DBLP:conf/nips/Dziugaite018}
Gintare~Karolina Dziugaite and Daniel~M. Roy.
\newblock Data-dependent {PAC-Bayes} priors via differential privacy.
\newblock In Samy Bengio, Hanna~M. Wallach, Hugo Larochelle, Kristen Grauman,
  Nicol{\`{o}} Cesa{-}Bianchi, and Roman Garnett, editors, \emph{Advances in
  Neural Information Processing Systems 31: Annual Conference on Neural
  Information Processing Systems 2018, NeurIPS 2018, December 3-8, 2018,
  Montr{\'{e}}al, Canada}, pages 8440--8450, 2018.
\newblock URL
  \url{https://proceedings.neurips.cc/paper/2018/hash/9a0ee0a9e7a42d2d69b8f86b3a0756b1-Abstract.html}.

\bibitem[Dziugaite et~al.(2020)Dziugaite, Drouin, Neal, Rajkumar, Caballero,
  Wang, Mitliagkas, and Roy]{DBLP:conf/nips/DziugaiteDNRCWM20}
Gintare~Karolina Dziugaite, Alexandre Drouin, Brady Neal, Nitarshan Rajkumar,
  Ethan Caballero, Linbo Wang, Ioannis Mitliagkas, and Daniel~M. Roy.
\newblock In search of robust measures of generalization.
\newblock In Hugo Larochelle, Marc'Aurelio Ranzato, Raia Hadsell,
  Maria{-}Florina Balcan, and Hsuan{-}Tien Lin, editors, \emph{Advances in
  Neural Information Processing Systems 33: Annual Conference on Neural
  Information Processing Systems 2020, NeurIPS 2020, December 6-12, 2020,
  virtual}, 2020.
\newblock URL
  \url{https://proceedings.neurips.cc/paper/2020/hash/86d7c8a08b4aaa1bc7c599473f5dddda-Abstract.html}.

\bibitem[Dziugaite et~al.(2021)Dziugaite, Hsu, Gharbieh, Arpino, and
  Roy]{DBLP:conf/aistats/DziugaiteHGA021}
Gintare~Karolina Dziugaite, Kyle Hsu, Waseem Gharbieh, Gabriel Arpino, and
  Daniel Roy.
\newblock On the role of data in pac-bayes.
\newblock In Arindam Banerjee and Kenji Fukumizu, editors, \emph{The 24th
  International Conference on Artificial Intelligence and Statistics, {AISTATS}
  2021, April 13-15, 2021, Virtual Event}, volume 130 of \emph{Proceedings of
  Machine Learning Research}, pages 604--612. {PMLR}, 2021.
\newblock URL
  \url{http://proceedings.mlr.press/v130/karolina-dziugaite21a.html}.

\bibitem[Foret et~al.(2021)Foret, Kleiner, Mobahi, and
  Neyshabur]{DBLP:conf/iclr/ForetKMN21}
Pierre Foret, Ariel Kleiner, Hossein Mobahi, and Behnam Neyshabur.
\newblock Sharpness-aware minimization for efficiently improving
  generalization.
\newblock In \emph{9th International Conference on Learning Representations,
  {ICLR} 2021, Virtual Event, Austria, May 3-7, 2021}. OpenReview.net, 2021.
\newblock URL \url{https://openreview.net/forum?id=6Tm1mposlrM}.

\bibitem[Gao and Zhou(2013)]{DBLP:journals/ai/GaoZ13a}
Wei Gao and Zhi{-}Hua Zhou.
\newblock On the doubt about margin explanation of boosting.
\newblock \emph{Artif. Intell.}, 203:\penalty0 1--18, 2013.
\newblock \doi{10.1016/j.artint.2013.07.002}.
\newblock URL \url{https://doi.org/10.1016/j.artint.2013.07.002}.

\bibitem[Germain et~al.(2009)Germain, Lacasse, Laviolette, and
  Marchand]{germainPACBayesianLearningLinear2009}
Pascal Germain, Alexandre Lacasse, Fran\c{c}ois Laviolette, and Mario Marchand.
\newblock {{PAC}}-{{Bayesian}} learning of linear classifiers.
\newblock In \emph{Proceedings of the 26th {{Annual International Conference}}
  on {{Machine Learning}} - {{ICML}} '09}, pages 1--8, {Montreal, Quebec,
  Canada}, 2009. {ACM Press}.
\newblock ISBN 978-1-60558-516-1.
\newblock \doi{10.1145/1553374.1553419}.

\bibitem[Gr{{\o}}nlund et~al.(2020)Gr{{\o}}nlund, Kamma, and
  Larsen]{DBLP:conf/icml/GronlundKL20}
Allan Gr{{\o}}nlund, Lior Kamma, and Kasper~Green Larsen.
\newblock Near-tight margin-based generalization bounds for support vector
  machines.
\newblock In \emph{Proceedings of the 37th International Conference on Machine
  Learning, {{ICML}} 2020, 13-18 July 2020, Virtual Event}, volume 119 of
  \emph{Proceedings of Machine Learning Research}, pages 3779--3788. {PMLR},
  2020.
\newblock URL \url{http://proceedings.mlr.press/v119/gronlund20a.html}.

\bibitem[Guedj(2019)]{guedj2019primer}
Benjamin Guedj.
\newblock {A Primer on PAC-Bayesian Learning}.
\newblock In \emph{Proceedings of the second congress of the French
  Mathematical Society}, 2019.
\newblock URL \url{https://arxiv.org/abs/1901.05353}.

\bibitem[Guedj and Alquier(2013)]{guedj2013}
Benjamin Guedj and Pierre Alquier.
\newblock {PAC-Bayesian estimation and prediction in sparse additive models}.
\newblock \emph{Electron. J. Statist.}, 7:\penalty0 264--291, 2013.
\newblock \doi{10.1214/13-EJS771}.
\newblock URL \url{https://doi.org/10.1214/13-EJS771}.

\bibitem[Hanneke and Kontorovich(2021)]{DBLP:conf/alt/HannekeK21}
Steve Hanneke and Aryeh Kontorovich.
\newblock Stable sample compression schemes: New applications and an optimal
  {SVM} margin bound.
\newblock In Vitaly Feldman, Katrina Ligett, and Sivan Sabato, editors,
  \emph{Algorithmic Learning Theory, 16-19 March 2021, Virtual Conference,
  Worldwide}, volume 132 of \emph{Proceedings of Machine Learning Research},
  pages 697--721. {PMLR}, 2021.
\newblock URL \url{http://proceedings.mlr.press/v132/hanneke21a.html}.

\bibitem[Jiang et~al.(2020)Jiang, Neyshabur, Mobahi, Krishnan, and
  Bengio]{DBLP:conf/iclr/JiangNMKB20}
Yiding Jiang, Behnam Neyshabur, Hossein Mobahi, Dilip Krishnan, and Samy
  Bengio.
\newblock Fantastic generalization measures and where to find them.
\newblock In \emph{8th International Conference on Learning Representations,
  {ICLR} 2020, Addis Ababa, Ethiopia, April 26-30, 2020}. OpenReview.net, 2020.
\newblock URL \url{https://openreview.net/forum?id=SJgIPJBFvH}.

\bibitem[Kristiadi et~al.(2020)Kristiadi, Hein, and
  Hennig]{DBLP:conf/icml/Kristiadi0H20}
Agustinus Kristiadi, Matthias Hein, and Philipp Hennig.
\newblock Being bayesian, even just a bit, fixes overconfidence in relu
  networks.
\newblock In \emph{Proceedings of the 37th International Conference on Machine
  Learning, {ICML} 2020, 13-18 July 2020, Virtual Event}, volume 119 of
  \emph{Proceedings of Machine Learning Research}, pages 5436--5446. {PMLR},
  2020.
\newblock URL \url{http://proceedings.mlr.press/v119/kristiadi20a.html}.

\bibitem[Langford and Seeger(2001)]{langford01boundsfor}
John Langford and Matthias Seeger.
\newblock Bounds for averaging classifiers.
\newblock 2001.
\newblock URL
  \url{http://www.cs.cmu.edu/~jcl/papers/averaging/averaging_tech.pdf}.

\bibitem[Langford and Shawe-Taylor(2003)]{langford2003pac}
John Langford and John Shawe-Taylor.
\newblock {{PAC}}-{{Bayes}} \& margins.
\newblock In \emph{Advances in Neural Information Processing Systems}, pages
  439--446, 2003.

\bibitem[LeCun et~al.(2010)LeCun, Cortes, and Burges]{lecun2010mnist}
Yann LeCun, Corinna Cortes, and CJ~Burges.
\newblock Mnist handwritten digit database.
\newblock \emph{ATT Labs [Online]. Available:
  http://yann.lecun.com/exdb/mnist}, 2, 2010.

\bibitem[Letarte et~al.(2019)Letarte, Germain, Guedj, and
  Laviolette]{NIPS2019_8911}
Ga\"el Letarte, Pascal Germain, Benjamin Guedj, and Francois Laviolette.
\newblock Dichotomize and generalize: {{PAC}}-{{Bayesian}} binary activated
  deep neural networks.
\newblock In H.~Wallach, H.~Larochelle, A.~Beygelzimer, F.~dAlch\'e Buc,
  E.~Fox, and R.~Garnett, editors, \emph{Advances in Neural Information
  Processing Systems 32}, pages 6872--6882. {Curran Associates, Inc.}, 2019.

\bibitem[Maurer(2004)]{DBLP:journals/corr/cs-LG-0411099}
Andreas Maurer.
\newblock A note on the {PAC-Bayesian} theorem.
\newblock \emph{CoRR}, cs.LG/0411099, 2004.
\newblock URL \url{http://arxiv.org/abs/cs.LG/0411099}.

\bibitem[McAllester(1998)]{McAllester1998}
David~A McAllester.
\newblock {Some PAC-Bayesian theorems}.
\newblock In \emph{Proceedings of the eleventh annual conference on
  Computational Learning Theory}, pages 230--234. ACM, 1998.

\bibitem[McAllester(1999)]{McAllester1999}
David~A McAllester.
\newblock {PAC-Bayesian model averaging}.
\newblock In \emph{Proceedings of the twelfth annual conference on
  Computational Learning Theory}, pages 164--170. ACM, 1999.

\bibitem[McAllester(2003)]{DBLP:conf/colt/McAllester03}
David~A. McAllester.
\newblock Simplified {{PAC}}-{{Bayesian}} margin bounds.
\newblock In Bernhard Sch\"olkopf and Manfred~K. Warmuth, editors,
  \emph{Computational Learning Theory and Kernel Machines, 16th Annual
  Conference on Computational Learning Theory and 7th Kernel Workshop,
  {{COLT}}/{{Kernel}} 2003, Washington, {{DC}}, {{USA}}, August 24-27, 2003,
  Proceedings}, volume 2777 of \emph{Lecture Notes in Computer Science}, pages
  203--215. {Springer}, 2003.
\newblock \doi{10.1007/978-3-540-45167-9\\_16}.

\bibitem[Nagarajan and Kolter(2019)]{DBLP:conf/nips/NagarajanK19}
Vaishnavh Nagarajan and J.~Zico Kolter.
\newblock Uniform convergence may be unable to explain generalization in deep
  learning.
\newblock In Hanna~M. Wallach, Hugo Larochelle, Alina Beygelzimer, Florence
  d'Alch{\'{e}}{-}Buc, Emily~B. Fox, and Roman Garnett, editors, \emph{Advances
  in Neural Information Processing Systems 32: Annual Conference on Neural
  Information Processing Systems 2019, NeurIPS 2019, December 8-14, 2019,
  Vancouver, BC, Canada}, pages 11611--11622, 2019.
\newblock URL
  \url{https://proceedings.neurips.cc/paper/2019/hash/05e97c207235d63ceb1db43c60db7bbb-Abstract.html}.

\bibitem[Neyshabur et~al.(2018)Neyshabur, Bhojanapalli, and
  Srebro]{DBLP:conf/iclr/NeyshaburBS18}
Behnam Neyshabur, Srinadh Bhojanapalli, and Nathan Srebro.
\newblock A {PAC-Bayesian} approach to spectrally-normalized margin bounds for
  neural networks.
\newblock In \emph{6th International Conference on Learning Representations,
  {ICLR} 2018, Vancouver, BC, Canada, April 30 - May 3, 2018, Conference Track
  Proceedings}. OpenReview.net, 2018.
\newblock URL \url{https://openreview.net/forum?id=Skz\_WfbCZ}.

\bibitem[Neyshabur et~al.(2019)Neyshabur, Li, Bhojanapalli, LeCun, and
  Srebro]{neyshabur2018the}
Behnam Neyshabur, Zhiyuan Li, Srinadh Bhojanapalli, Yann LeCun, and Nathan
  Srebro.
\newblock The role of over-parametrization in generalization of neural
  networks.
\newblock In \emph{International Conference on Learning Representations}, 2019.
\newblock URL \url{https://openreview.net/forum?id=BygfghAcYX}.

\bibitem[Novikoff(1962)]{novikoff62convergence}
A.~B. Novikoff.
\newblock On convergence proofs on perceptrons.
\newblock In \emph{Proceedings of the Symposium on the Mathematical Theory of
  Automata}, volume~12, pages 615--622, New York, NY, USA, 1962. Polytechnic
  Institute of Brooklyn.

\bibitem[Parrado{-}Hern{\'{a}}ndez et~al.(2012)Parrado{-}Hern{\'{a}}ndez,
  Ambroladze, Shawe{-}Taylor, and
  Sun]{DBLP:journals/jmlr/Parrado-HernandezASS12}
Emilio Parrado{-}Hern{\'{a}}ndez, Amiran Ambroladze, John Shawe{-}Taylor, and
  Shiliang Sun.
\newblock Pac-bayes bounds with data dependent priors.
\newblock \emph{J. Mach. Learn. Res.}, 13:\penalty0 3507--3531, 2012.
\newblock URL \url{http://dl.acm.org/citation.cfm?id=2503353}.

\bibitem[Schapire et~al.(1998)Schapire, Freund, Bartlett, and
  Lee]{schapireBoostingMarginNew1998}
Robert~E. Schapire, Yoav Freund, Peter Bartlett, and Wee~Sun Lee.
\newblock Boosting the margin: A new explanation for the effectiveness of
  voting methods.
\newblock \emph{The Annals of Statistics}, 26\penalty0 (5):\penalty0
  1651--1686, October 1998.
\newblock \doi{10.1214/aos/1024691352}.

\bibitem[Seeger et~al.(2001)Seeger, Langford, and Megiddo]{seeger2001improved}
Matthias Seeger, John Langford, and Nimrod Megiddo.
\newblock An improved predictive accuracy bound for averaging classifiers.
\newblock In \emph{Proceedings of the 18th International Conference on Machine
  Learning}, number CONF, pages 290--297, 2001.

\bibitem[Shawe-Taylor and Williamson(1997)]{STW1997}
J.~Shawe-Taylor and R.~C. Williamson.
\newblock {A PAC analysis of a Bayes estimator}.
\newblock In \emph{Proceedings of the 10th annual conference on Computational
  Learning Theory}, pages 2--9. ACM, 1997.

\bibitem[Tropp(2012)]{DBLP:journals/focm/Tropp12}
Joel~A. Tropp.
\newblock User-friendly tail bounds for sums of random matrices.
\newblock \emph{Found. Comput. Math.}, 12\penalty0 (4):\penalty0 389--434,
  2012.
\newblock \doi{10.1007/s10208-011-9099-z}.
\newblock URL \url{https://doi.org/10.1007/s10208-011-9099-z}.

\bibitem[Zhang et~al.(2021)Zhang, Bengio, Hardt, Recht, and
  Vinyals]{DBLP:journals/cacm/ZhangBHRV21}
Chiyuan Zhang, Samy Bengio, Moritz Hardt, Benjamin Recht, and Oriol Vinyals.
\newblock Understanding deep learning (still) requires rethinking
  generalization.
\newblock \emph{Commun. {ACM}}, 64\penalty0 (3):\penalty0 107--115, 2021.
\newblock \doi{10.1145/3446776}.
\newblock URL \url{https://doi.org/10.1145/3446776}.

\bibitem[Zhou et~al.(2019)Zhou, Veitch, Austern, Adams, and
  Orbanz]{DBLP:conf/iclr/ZhouVAAO19}
Wenda Zhou, Victor Veitch, Morgane Austern, Ryan~P. Adams, and Peter Orbanz.
\newblock Non-vacuous generalization bounds at the {ImageNet} scale: a
  {PAC-Bayesian} compression approach.
\newblock In \emph{7th International Conference on Learning Representations,
  {ICLR} 2019, New Orleans, LA, USA, May 6-9, 2019}. OpenReview.net, 2019.
\newblock URL \url{https://openreview.net/forum?id=BJgqqsAct7}.

\end{thebibliography}
